\documentclass[]{fairmeta}
\usepackage{xcolor}
\usepackage{colortbl}
\usepackage{amsfonts}
\usepackage{xspace}
\usepackage{graphicx}
\usepackage{tabularx}
\usepackage{pifont}
\usepackage{overpic}
\usepackage{makecell}
\usepackage{tabularx}

\usepackage{arydshln}
\usepackage{wrapfig}
\usepackage{subcaption}
\usepackage{booktabs}
\usepackage{amssymb}
\usepackage{xifthen}
\usepackage{wrapfig}
\usepackage{makecell}

\usepackage{multirow}
\usepackage{booktabs} 
\usepackage{amsmath}
\usepackage[normalem]{ulem}
\usepackage{xspace}
\usepackage{pifont}
\usepackage[inkscapelatex=false]{svg}

\title{Xray-Visual Models: Scaling Vision models on Industry Scale Data}

\author[1]{Shlok Mishra}
\author[1]{Tsung-Yu Lin}
\author[1]{Linda Wang}
\author[1]{Hongli Xu}
\author[1]{Yimin Liu}
\author[1]{Michael Hsu}
\author[1]{Chaitanya Ahuja}
\author[1]{Hao Yuan}
\author[1]{Jianpeng Cheng}
\author[1]{Hong-You Chen}
\author[1]{Haoyuan Xu}
\author[1,2]{Chao Li}
\author[1]{Sreya Dutta Roy}
\author[1]{Abhijeet Awasthi}
\author[1]{Jihye Moon}
\author[1]{Don Husa}
\author[1]{Michael Ge}
\author[1]{Sumedha Singla}
\author[1]{Arkabandhu Chowdhury}
\author[1]{Phong Dingh} 
\author[1]{Satya Narayan Shukla}
\author[1]{Yonghuan Yang}
\author[1,3]{David Jacobs}
\author[1]{Qi Guo}
\author[1]{Jun Xiao}
\author[1]{Xiangjun Fan}
\author[1]{Aashu Singh}
\affiliation[1]{Meta-AI}
\affiliation[2]{MIT (work done while at Meta AI)}
\affiliation[3]{University of Maryland (work done while at Meta AI)}

\abstract{
We present Xray-Visual, a unified vision model architecture for large-scale image and video understanding trained on industry-scale social media data. Our model leverages over 15 billion curated image-text pairs and 10 billion video-hashtag pairs from Facebook and Instagram, employing robust data curation pipelines that incorporate balancing and noise suppression strategies to maximize semantic diversity while minimizing label noise. We introduce a three-stage training pipeline that combines self-supervised MAE, semi-supervised hashtag classification, and CLIP-style contrastive learning to jointly optimize image and video modalities. Our architecture builds on a Vision Transformer backbone enhanced with efficient token reorganization (EViT) for improved computational efficiency. Extensive experiments demonstrate that Xray-Visual achieves state-of-the-art performance across diverse benchmarks, including ImageNet for image classification, Kinetics and HMDB51 for video understanding, and MSCOCO for cross-modal retrieval. The model exhibits strong robustness to domain shift and adversarial perturbations. We further demonstrate that integrating large language models as text encoders (LLM2CLIP) significantly enhances retrieval performance and generalization capabilities, particularly in real-world  environments. Xray-Visual establishes new benchmarks for scalable, multimodal vision models, while maintaining superior accuracy and computational efficiency.
}

\date{\today}
\correspondence{\email{shlokmishra@meta.com}}

\begin{document}

\maketitle

\section{Introduction}

\begin{figure}[ht]
    \centering
    \begin{subfigure}[b]{0.48\textwidth}
        \centering
        \includegraphics[width=\textwidth]{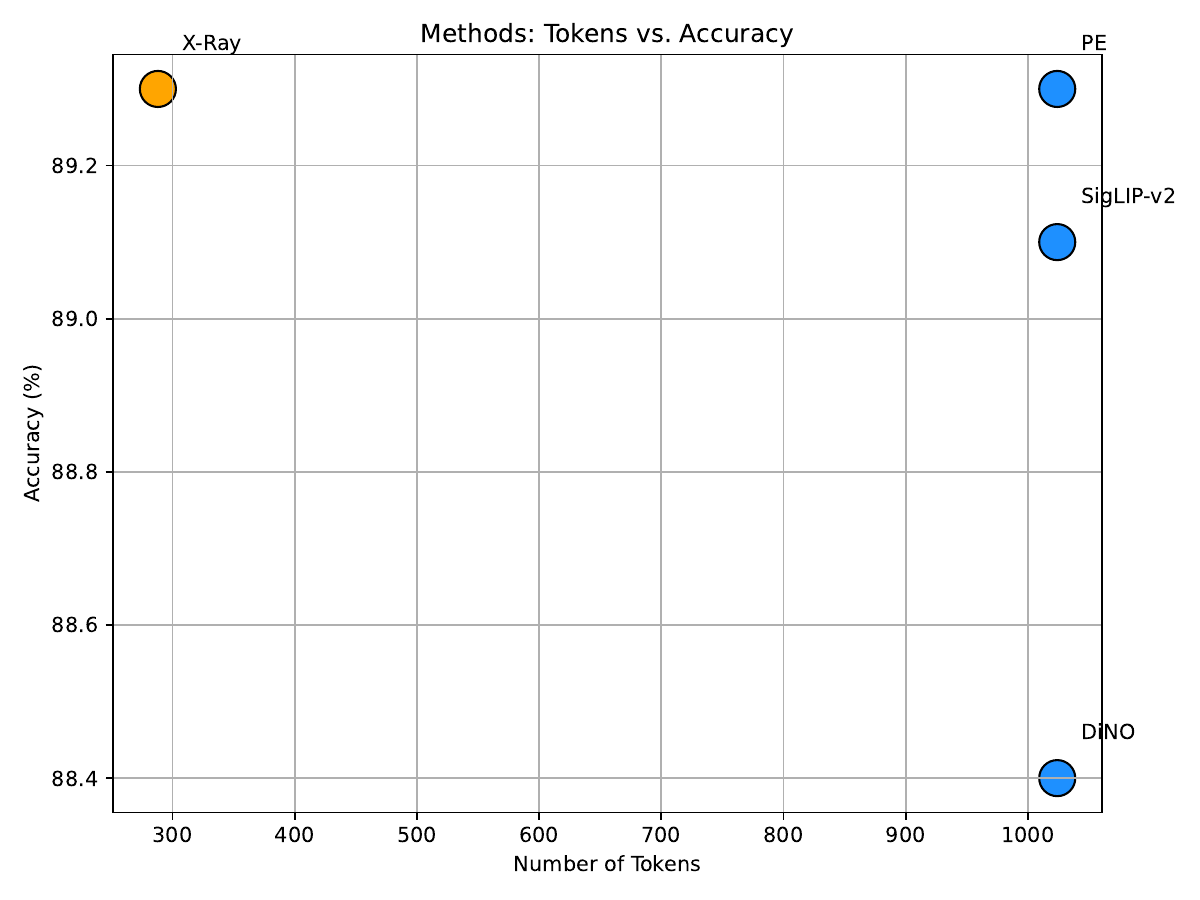}
        \caption{ImageNet Accuracy vs. Tokens}
        \label{fig:imagenet}
    \end{subfigure}
    \hfill
    \begin{subfigure}[b]{0.48\textwidth}
        \centering
        \includegraphics[width=\textwidth]{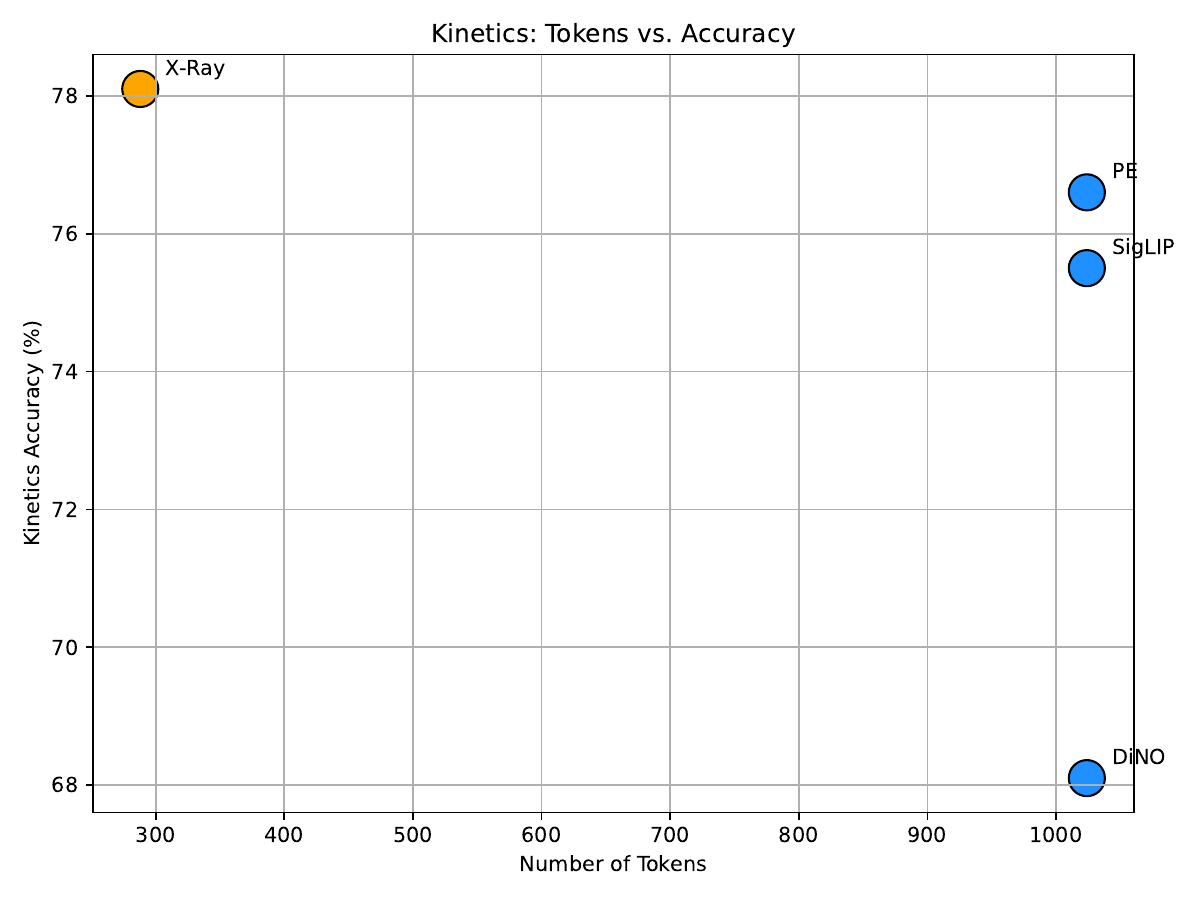}
        \caption{Kinetics Accuracy vs. Tokens}
        \label{fig:kinetics}
    \end{subfigure}
    \caption{XRay achieves 89.3\% Top-1 accuracy with 336px and 288 tokens, vs. baselines at 448px/1024 tokens. This is a 71.9\% token reduction, 43.75\% pixel area reduction, and 84.2\% reduction in the product proxy (~6.3× lower combined cost).}
    \label{fig:comparison}
\end{figure}

Recent years have witnessed extraordinary advances in computer vision, particularly in image and video understanding, driven by innovations in deep learning architectures and large-scale pre-training objectives \citep{he2015deepresiduallearningimage, simclr, he2020momentum, radford2021learning, SLIP, siglip, kay2017kineticshumanactionvideo}. Despite these achievements, vision models still lag behind large language models (LLMs) in terms of data scale and generalization \citep{llama, gpt, gpt3}. The success of LLMs is largely attributed to the abundance of human-generated text available for training, whereas visual data at comparable scale remains challenging to collect and curate. This disparity has limited the progress of vision pre-training, especially in achieving robust performance across diverse, real-world scenarios.

\begin{figure}[ht]
    \centering
    \begin{subfigure}[b]{0.55\textwidth}
        \centering
        \includegraphics[width=\textwidth]{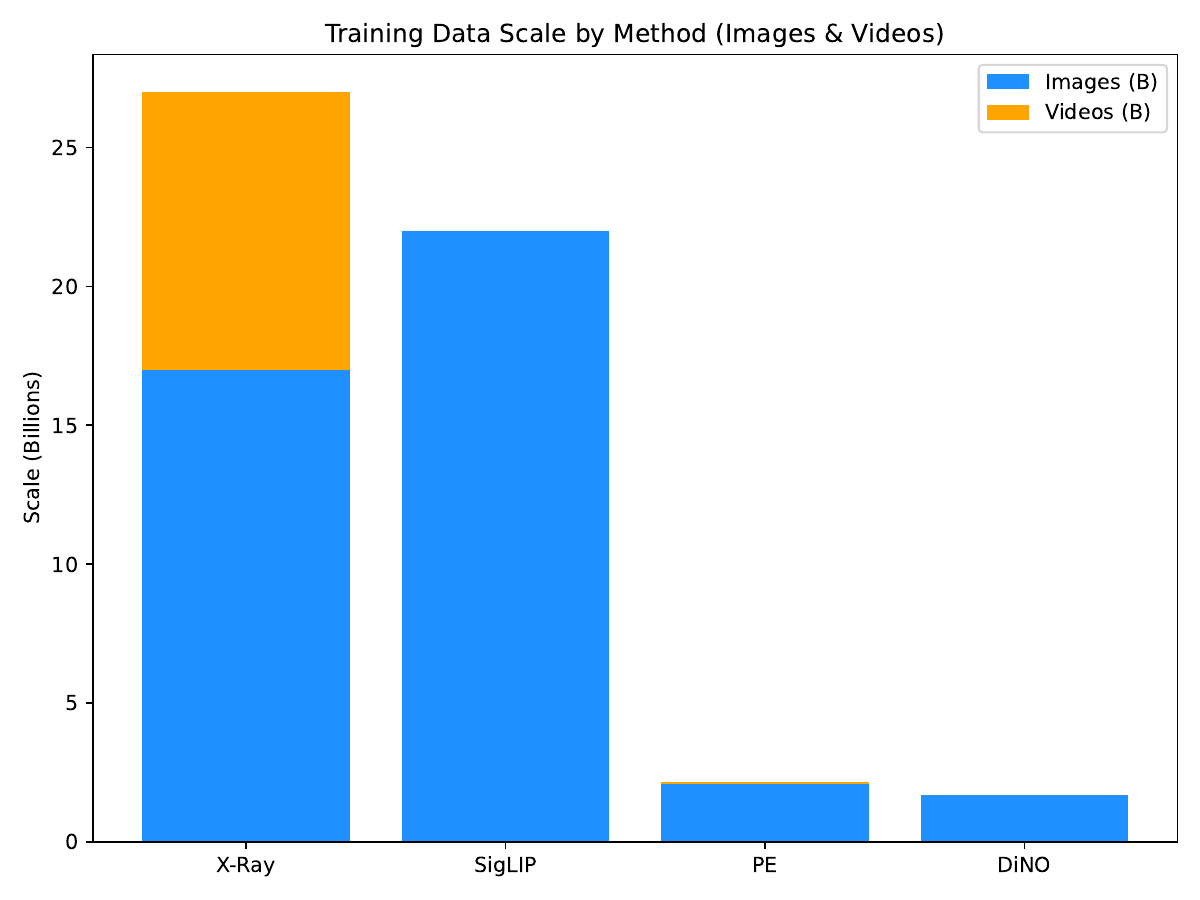}
        \caption{Training data scale for each method (images and videos).}
        \label{fig:images_videos}
    \end{subfigure}
    \hfill
    \begin{subfigure}[b]{0.35\textwidth}
        \centering
        \includegraphics[width=\textwidth]{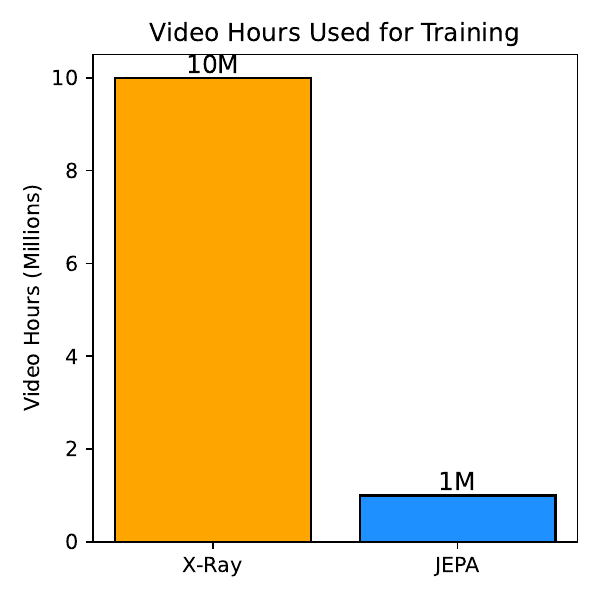}
        \caption{Video hours used for training (X-Ray vs. JEPA). }
        \label{fig:video_hours}
    \end{subfigure}
    \caption{\textbf{Comparison of training data scale across vision models.} (a) XrayVisual leverages the largest curated dataset for vision encoder training to date, (b) We utilize 10$\times$ more video data than state-of-the-art world models such as V-JEPA~\cite{assran2025vjepa2selfsupervisedvideo}.}
    \label{fig:data_scale_comparison}
\end{figure}

To address this gap, we introduce a new family of unified vision models trained on billions of social media images and videos, setting new state-of-the-art (SOTA) results on both academic benchmarks and real-world tasks. Our approach is grounded in a comprehensive data curation pipeline that transforms raw, noisy corpora—initially exceeding 100 billion images and videos—into high-quality, semantically rich supervision. The ViSE image dataset comprises of over 10 billion image-text pairs sourced from public Facebook and Instagram posts, with captions and hashtags meticulously cleaned, filtered for English, and balanced to ensure broad semantic coverage. For video understanding, we curate 10 billion video-hashtag pairs from 24 billion Instagram posts, employing canonicalization and long-tail resampling to guarantee adequate representation of rare visual concepts. In Fig \ref{fig:data_scale_comparison} we show that we train on 26 Billion images and videos which is the largest amount of pre-training in pubically avaliable vision encoder pre-training. For videos we train on 10 billion videos which is close to 10m hours of video data, which is 10x more than world models like JEPA \cite{assran2025vjepa2selfsupervisedvideo} as shown in Fig \ref{fig:video_hours}. Supervision quality is further enhanced by generating synthetic captions for both images and videos using \cite{grattafiori2024llama3herdmodels}, followed by text only LLM-based rewriting \cite{fan2023improvingcliptraininglanguage} to increase diversity and reduce repetition \citep{llama}.

Our models employ a unified architecture that jointly processes images and videos, enabling robust multimodal representation learning. This is done by extending the Vision Transformer (ViT) backbone \citep{vit} with 3D tokenization \cite{tran2015learningspatiotemporalfeatures3d}, allowing seamless handling of multiple modalities within a single framework. Our primary vision backbone follows EViT \cite{evit}; which focuses on efficiency while maintaining high performance. EViT  \citep{evit},  reorganizes and drop inattentive tokens, supporting higher input resolutions with minimal computational overhead. Register tokens are appended to the transformer sequence \citep{Registers}, enhancing the model’s ability to capture outlier features and boosting video performance.
The training pipeline consists of three key stages: (1) self-supervised masked autoencoding (MAE) \citep{mae} for learning visual representations from masked patches in images and videos; (2) semi-supervised hashtag classification leveraging large-scale hashtag-labeled datasets; and (3) CLIP-style contrastive learning \citep{radford2021learning} using curated image and video caption pairs to align visual and textual representations. Additional objectives such as denoising \citep{chen2024deconstructing} and SLIP (Self-supervision meets Language-Image Pre-training) \citep{SLIP} further strengthen the learned features. Notably, we replace standard text encoders with LLaMA-1b \citep{llama}, which proves critical for high-performance retrieval on real world datasets and internal metrics.

Empirically, our pure image models achieve SOTA 89.3\% accuracy on ImageNet \cite{imagenet}, while the unified image-video architecture attains 88.1\% on ImageNet and 78.1\% on Kinetics — accomplished with lower (336)input resolutions and only 1/4th (288 tokens) the tokens compared to other leading vision models like PE \cite{bolya2025PerceptionEncoder}, SigLIP \citep{siglip2,siglip}, which use 448 resolution and 1024 tokens as can be seen in Fig \ref{fig:imagenet} and Fig \ref{fig:kinetics}.  We also observe that our joint image-video model trained on both image and video is better for video understanding then pure video only models; due to the synergy of image and video and cross domain information. It's also almost SOTA on image bechmarks and achieves 88.1\% only 1.2\% below pure image models. We also set new SOTA benchmarks on MS-COCO \cite{coco} image-to-text and text-to-image retrieval. Across a wide range of data scales, our models demonstrate strong generalization and robust out-of-distribution (OOD) performance. Most importantly, we show that existing vision encoders suffer significant performance drops on real-world data distributions, whereas our LLM text encoder based models consistently deliver best-in-class results on complex, production-scale retrieval tasks.


\paragraph{Overview of Contributions}
Our main contributions are summarized below.
\begin{itemize}
\item \textbf{Data collection}
     \begin{itemize}
    \item We show a systematic study of how to train vision models on industry scale data. We developed a comprehensive data curation pipeline for large-scale vision model training, starting from over 100 billion raw social media images and videos.
\item We curate two large-scale datasets: (1) ViSE image dataset with over 15 billion image-text pairs from public Facebook and Instagram posts, applying rigorous cleaning (removal of URLs, emojis, user tags, and non-semantic content), language detection for English captions, and WordNet synset-based resampling to balance semantic diversity across the long-tail distribution, and (2) a video dataset with 10 billion video-hashtag pairs from 24 billion Instagram posts, with canonicalized hashtags and resampling to ensure robust representation of both common and rare visual concepts.
    \end{itemize}
     \item \textbf{Model Architecture and Training}
     \begin{itemize}
    \item We proposed a unified model architecture that begins with self-supervised masked autoencoding (MAE) for initial representation learning, followed by fine-tuning with hashtag classification, and culminating in CLIP-style contrastive training using curated image and video caption pairs.
    \item  Incorporated denoising loss and added LLM's as the text encoder during training to further enhance scaling behavior and robustness, demonstrating that larger models and datasets yield consistent performance improvements and LLM's help in real world generalization. 
\item  Designed a joint image-video architecture based on Vision Transformers, enabling simultaneous learning from both modalities. This joint approach significantly improves video representations and remains highly competitive for image tasks, eliminating the need for separate models.\end{itemize}
    \item \textbf{Efficiency and Performance}
     \begin{itemize}
    \item  We achieved state-of-the-art results on major benchmarks including ImageNet, Kinetics, MS-COCO, and various out-of-distribution (OOD) datasets, validating the model’s generalization and robustness. 
    \item Demonstrated that our model operates at a 336 input resolution while utilizing only 25\% of the tokens (288 tokens) compared to Perception Encoder (PE) \cite{bolya2025PerceptionEncoder}, DiNO (1024 tokens) \cite{siméoni2025dinov3}, SigLIP \cite{siglip2} resulting in a 4x increase in queries per second (QPS) and substantial computational efficiency. 
\item Due to training on large scale datasets, Despite the reduced token usage and higher efficiency, our model consistently outperforms or matches the best existing models on both academic and real-world tasks.\end{itemize}
\end{itemize}

\section{Data} In this section we discuss the details of data curation. Our training pipeline is in three stages, we begin with MAE followed by finetuning and in third stage we use CLIP training. We use different datasets in different stages.  We begin with describing image data for all the three stages followed by details of video data.

\subsection{Image Dataset - ViSE} 
We assembled a large-scale, high-quality dataset of image-text pairs from Facebook and Instagram called ViSE. Raw social media captions typically contain extraneous content such as URLs, email addresses, emojis, user tags, and non-semantic characters. We applied rigorous post-processing to remove these elements and retain only meaningful text, including removing the pound sign from hashtags while preserving the hashtag text.

\paragraph{\textbf{Semantic balancing}}
A key challenge in web-scale data curation is the imbalanced distribution of visual concepts~\cite{clip}. We leveraged WordNet synsets to identify visually relevant concepts and deduplicate synonyms. To address the long-tail distribution, we resampled the dataset by undersampling frequent (head) concepts and oversampling rare (tail) concepts. Specifically, we segmented captions into sentences using a custom UDF, associating each image with multiple sentences as distinct pairs. We computed unigram frequencies (excluding English stopwords) and resampled accordingly. As shown in Table~\ref{tab:vise_sampling_accuracy}, this balancing strategy yielded a 7.3\% improvement in ImageNet linear probe accuracy for ViT-B models trained from scratch for 700K iterations.

\begin{table}[h]
\centering
\begin{tabular}{|l|c|}
\hline
\textbf{Method} & \textbf{Accuracy} \\
\hline
ViSE without resampling & 65.4\% \\
ViSE with resampling & 72.7\% \\
\hline
\end{tabular}
\caption{Impact of semantic resampling on ImageNet linear classification accuracy.}
\label{tab:vise_sampling_accuracy}
\end{table}

\paragraph{\textbf{Noise suppression}}
To mitigate noise inherent in web-scale data, we applied similarity-based filtering using a pre-trained MetaCLIP~\cite{metaclip} model to score image-text alignment. We experimented with various similarity thresholds and additionally incorporated label smoothing into the contrastive loss. Table~\ref{tab:ablation_filtering} shows that filtering with a 0.25 similarity threshold combined with label smoothing achieved the best results, improving ImageNet linear classification accuracy from 71.7\% to 72.8\%. All models were trained for 200K iterations. The final ViSE dataset contains approximately 10 billion high-quality image-text pairs, making it one of the largest datasets used for CLIP-style training~\cite{LAOIN, metaclip, metaclip2}.

\begin{table}[h!]
\centering
\begin{tabular}{|c|c|c|}
\hline
\makecell{\textbf{Filtering}} & \makecell{\textbf{Method}} & \makecell{\textbf{Top-1} \\ \textbf{Accuracy (\%)}} \\
\hline
 & Resampling & 71.7 \\
\hline
\checkmark & + Filtering & 72.8 \\
\hline
\end{tabular}
\caption{Ablation study on ImageNet linear classification: impact of similarity-based filtering.}
\label{tab:ablation_filtering}
\end{table}

\subsection{Video Data}
To train high-quality multimodal representations, we curate a large-scale video-text dataset from Instagram posts called URU. Unlike user-provided captions for videos, which are often generic or noisy, hashtags typically contain more precise visual and semantic descriptors directly relevant to video content. We systematically process hashtags from billions of public Instagram posts containing videos, resulting in a curated collection of $\sim$5 billion video-hashtag pairs.

\paragraph{\textbf{Hashtag Processing and Canonicalization}} 
User-provided hashtags can be inherently noisy and irrelevant. To improve data quality and hashtag effectiveness, we follow the procedure of prior work \cite{Singh2022RevisitingWS} for images with additional optimizations to extract meaningful hashtags for training purposes. We leveraged WordNet synsets to identify hashtags related to meaningful objects. We then map semantically similar hashtags to canonical representations, creating $\sim$75k user-provided hashtag - canonical hashtag pairs. This canonicalization process reduces vocabulary size while improving semantic consistency and training stability. 

\paragraph{\textbf{Video-Hashtag Pair Construction}}
When any user-provided hashtag can be mapped to our user-provided - canonical hashtag pairs, videos - user-provided hashtag pair is retrieved to construct a new video - canonical hashtag pair. After addressing the long-tail distribution by downsampling frequent hashtags and oversampling rare hashtags, we generated $\sim$5 billion video - canonical hashtags pairs.

\paragraph{\textbf{Adapted Data for Training Strategy}}
As described in Section 4.2, we adopt a three-stage pre-training process. To adapt our training strategy, we utilized different types of hashtags as supervisory signals, following the findings in \cite{Ghadiyaram2019LargeScaleWP}. Importantly, our larger scale datasets are designed to promote general visual understanding capabilities, rather than overlapping with target datasets such as IG-Kinetics described in the same study. For the first-stage self-supervised MAE and the second-stage semi-supervised hashtag classification, we employed URU-Video-Noun-5B, which focuses exclusively on hashtags related to visual objects. To further enhance the model’s capacity for action understanding alongside object recognition, we expanded the hashtag set and construct URU-Video-Action-5B for the third-stage CLIP training.

\begin{table}[h]
\centering
\begin{tabular}{|c|c|c|}
\hline
\textbf{Dataset} & \textbf{Total \#labels(hashtags)} \\
\hline
IG-Kinetics & 359 \\
URU-Video-Noun-5B & 22013 \\
URU-Video-Action-5B & 31619 \\
\hline
\end{tabular}
\caption{Comparison of label scale across video-hashtag datasets. IG-Kinetics \cite{Ghadiyaram2019LargeScaleWP} uses 400 action labels from Kinetics as seed labels, resulting in 359 unique labels. In contrast, the URU-Video datasets leverage a broader vocabulary filtered from WordNet synsets, yielding significantly more labels and supporting richer visual and action concept coverage.}
\label{tab:dataset_scale}
\end{table}

\subsection{Synthetic Captions}

Synthetic data generation has become an effective strategy for training vision-language models~\cite{LLM_rewrite, Zhu2025InternVL3EA, gemma, llama}. We generated synthetic captions using an internal multimodal large language model (MMLLM) trained on Facebook and Instagram videos and reels.

Our experiments revealed a trade-off: while synthetic captions significantly improved retrieval performance compared to user-generated captions, they led to decreased video classification accuracy. We hypothesized that this degradation stemmed from quality issues in the generated captions, particularly repetitive phrasing. Analysis revealed approximately 3\% repetition our video captions. To fix this we propose to use LLM based rewrites which we mention in next section.

\subsubsection{LLM-Based Caption Refinement}

To address caption quality issues, we leveraged LLM-based rewriting~\cite{LLM_rewrite}, which has been shown to increase diversity in sentence structure and vocabulary while preserving semantic content. We employed Llama 3B~\cite{grattafiori2024llama3herdmodels} to refine the synthetic captions, using prompts such as ``rewrite this caption of a video vividly, and keep it less than thirty words.'' This approach effectively eliminated repetitive elements and produced more concise, descriptive captions.


\subsection{Other datasets}
We also use MetaCLIP \cite{metaclip2} dataset for pre-training. MetaCLIP dataset is very helpful specially for pre-training for captioning realted tasks. Whenver we are pre-training for MMLLM's we found that higher quality metaclip dataset had higher performance for downstream captioning tasks. However in general for internal embedding related tasks using MetaCLIP we saw regresssion on internal retrieval tasks.We also hashtag the dataset for images (\(\sim 5\)B images) and videos (\(\sim 5\)B videos) for second-stage pre-training expanding ~\cite{mahajan2018exploringlimitsweaklysupervised}.
\section{Model architecture}
Generally, computer vision models employ distinct architectures for image and video tasks \citep{simclr, dino, mae}. However, for large-scale industrial training, it is desirable to leverage both image and video datasets jointly to obtain richer representations. To this end, we adopt a co-training approach, training on both images and videos simultaneously. We observe that joint training consistently improves video performance while maintaining strong image performance. An additional benefit of a unified architecture is the reduced deployment overhead, as a single model can be used for both modalities, eliminating the need to maintain and deploy separate models for images and videos.

\subsection {Design Architecture}
The base backbone used in XrayVisual training is the Vision Transformer (ViT) architecture~\cite{vit}. To enable joint training on both images and videos, we conducted extensive ablation studies and selected a video ViT backbone with 3D tokenization for XrayVisual.
Special consideration is required when processing image inputs. Specifically, we repeat each image along the temporal dimension to match the kernel size of the 3D convolution, and apply zero-padding and unpadding to accommodate the positional embeddings. For video inputs, no such modification is necessary: the 3D convolution operates directly on the full video sequence without the need for zero-padding.

\begin{figure}[htbp]
    \centering
    \includegraphics[width=0.7\textwidth]{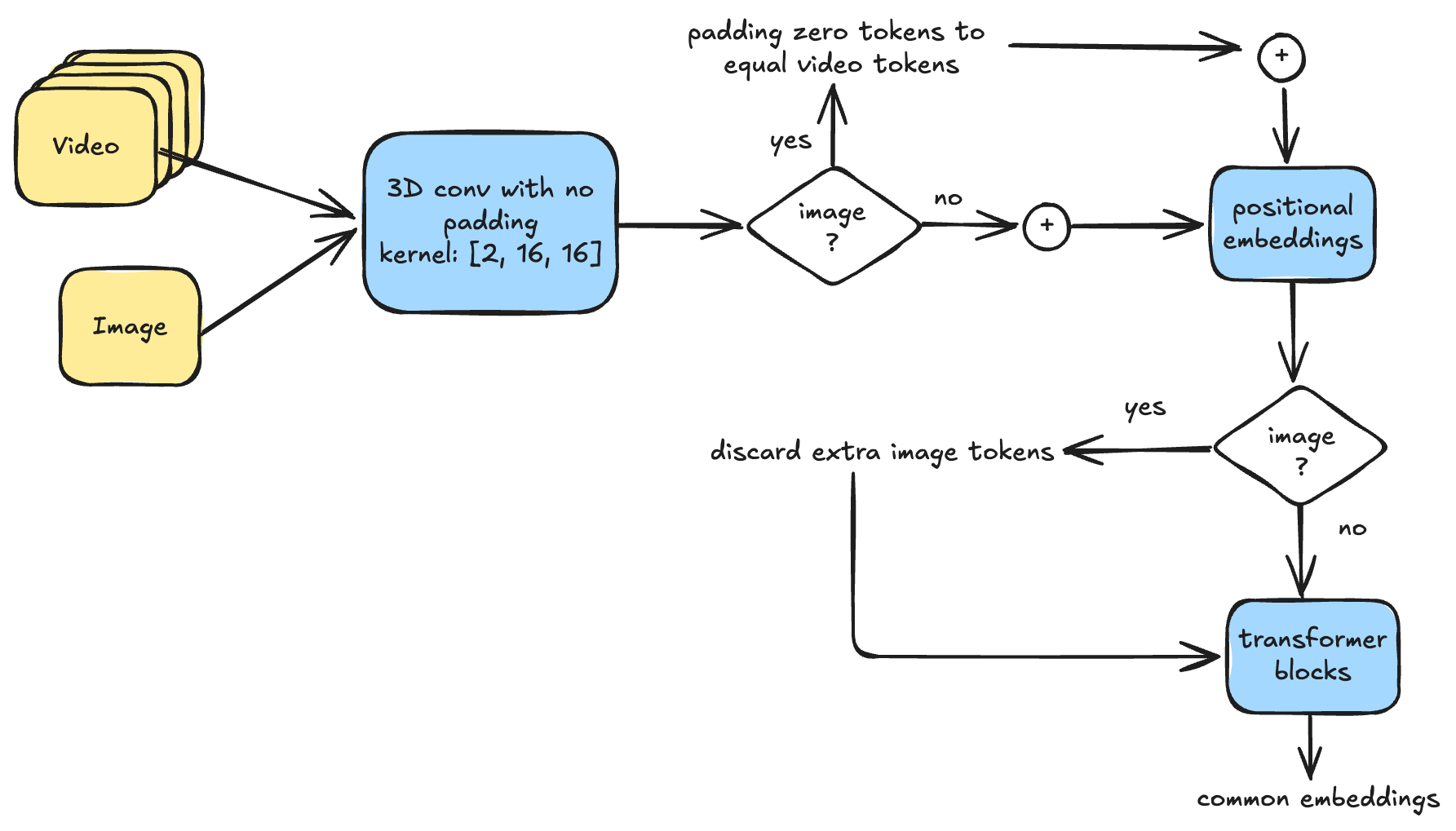}
     \caption{XRV Architecture: XrayVisual uses a Vision Transformer (ViT)~\cite{vit} backbone with 3D tokenization for joint image-video training. Images are repeated along the temporal dimension with zero-padding to match the 3D convolution kernel and positional embeddings, while videos are processed directly without modification. }
    \label{fig:xrv_architectuure}
\end{figure}

\subsection{EViT-2b}
For optimal performance, we employ larger 2 billion parameter models utilizing the EViT architecture~\cite{evit}.

\paragraph{\textbf{Efficient high-resolution processing}}
While higher input resolutions capture more detailed visual information and typically improve recognition accuracy \cite{vit}, they pose computational challenges for vision transformers. Specifically, increasing input resolution leads to quadratic growth in the number of tokens, substantially increasing computational cost. To mitigate this issue, we adopt the token reorganization technique from EViT~\cite{evit}, which selectively prunes inattentive tokens to improve efficiency while preserving accuracy.

The EViT method computes attention weights from the CLS token to all patch tokens, identifying those with low attention scores as inattentive. By tuning the keep rate hyperparameter, we can achieve a favorable trade-off between computational efficiency and task performance. Empirically, we find that removing inattentive tokens incurs only negligible accuracy degradation while enabling the model to process higher-resolution inputs and achieve superior overall performance.

We adapt this technique to image-text contrastive learning and observe approximately 0.5\% improvement over standard ViT models. Following the EViT protocol~\cite{evit}, we uniformly select three transformer layers for token reorganization. Table~\ref{tab:evit} presents our results, where throughput (images/sec) is measured on a single V100 GPU. Based on these experiments, we adopt 288$\times$288 resolution for ViT-B16 and 336$\times$336 resolution for models larger than ViT-H14. All final results are reported using a 2 billion parameter model with EViT.

\begin{table}[h!]
\centering
\begin{tabular}{|c|c|c|c|c|c|}
\hline
\textbf{Model} & \textbf{Resolution} & \textbf{Keep Rate} & \textbf{Throughput} & \textbf{LC (\%)} & \textbf{ZS (\%)} \\
\hline
ViT-B16  & 224 & 1.0 & 227.23 & 83.11 & 72.24 \\
ViT-B16  & 288 & 0.7 & 281.87 & 83.57 & 72.79 \\
ViT-B16  & 336 & 0.5 & 219.25 & 83.27 & 72.55 \\
ViT-H14  & 224 & 1.0 & 29.85  & 86.99 & 79.05 \\
ViT-H14  & 280 & 0.7 & 25.90  & 86.77 & 79.11 \\
ViT-H14  & 336 & 0.5 & 23.87  & 87.38 & 80.03 \\
\hline
\end{tabular}
\caption{Model Performance at Different Resolutions and Keep Rates. }
\label{tab:evit}
\end{table}
\section{Training}
In this section we describe our training methodology beginning with batch sampling Strategy, followed by our three stage training pipeline.
\subsection{Multi-Modal Batch Sampling Strategy}

We observe that the distribution of image and video data during joint training significantly impacts model performance. Prolonged consecutive exposure to a single modality can introduce bias toward that modality. Our experiments reveal that the critical factor is not the absolute quantity of data from each modality, but rather the frequency and uniformity with which the model encounters both modalities during training. While moderately imbalanced image-to-video ratios (up to 1:10) remain acceptable, it is essential to distribute batches from both modalities uniformly throughout each epoch, preventing extended periods of single-modality exposure.

In our training scheme, each batch contains data from only one modality. Through extensive experimentation, we identify the optimal strategy: sampling image or video batches according to a predefined probability that ensures both modalities are exhausted approximately simultaneously within each epoch. This approach uniformly distributes the smaller modality across the larger one, maintaining balanced exposure throughout training.

Formally, consider a scenario with $N_I$ image samples (batch size $B_I$) and $N_V$ video samples (batch size $B_V$). The sampling probabilities for image and video batches are computed as:
$$P_I : P_V = \frac{N_I}{B_I} : \frac{N_V}{B_V}$$

For example, with 5B image samples (batch size 64) and 2B video samples (batch size 32), the ratio becomes $(5/64):(2/32) = 5:4 = 0.56:0.44$. Thus, at each iteration, the model samples an image batch with probability 0.56 and a video batch with probability 0.44.

Table~\ref{tab:batch_sampling_strategies} demonstrates the importance of selecting an appropriate batch sampling strategy, evaluated using ViT-L on URU training data with an image-to-video ratio of 5B:500M.

\begin{table}[h]
\centering
\begin{tabular}{|p{7.5cm}|c|c|}
\hline
\textbf{Batch Sampling Strategy} & \textbf{ImageNet LP} & \textbf{Kinetics Top-1} \\
\hline
Strict alternation; new epoch starts after image modality finishes, resulting in image-only training after video modality exhaustion & 86.03\% & 71.2\% \\
\hline
Strict alternation; video modality repeats after exhaustion & 80.56\% & 73.3\% \\
\hline
Probabilistic sampling; both modalities finish epoch simultaneously (Ours) & 84.16\% & 74.1\% \\
\hline
\end{tabular}
\caption{Comparison of batch sampling strategies on ImageNet linear probing accuracy and Kinetics Top-1 accuracy.}
\label{tab:batch_sampling_strategies}
\end{table}

\subsection{Multi-Stage Training Strategy}

\begin{figure}[ht]
    \centering
    \includegraphics[width=0.75\columnwidth]{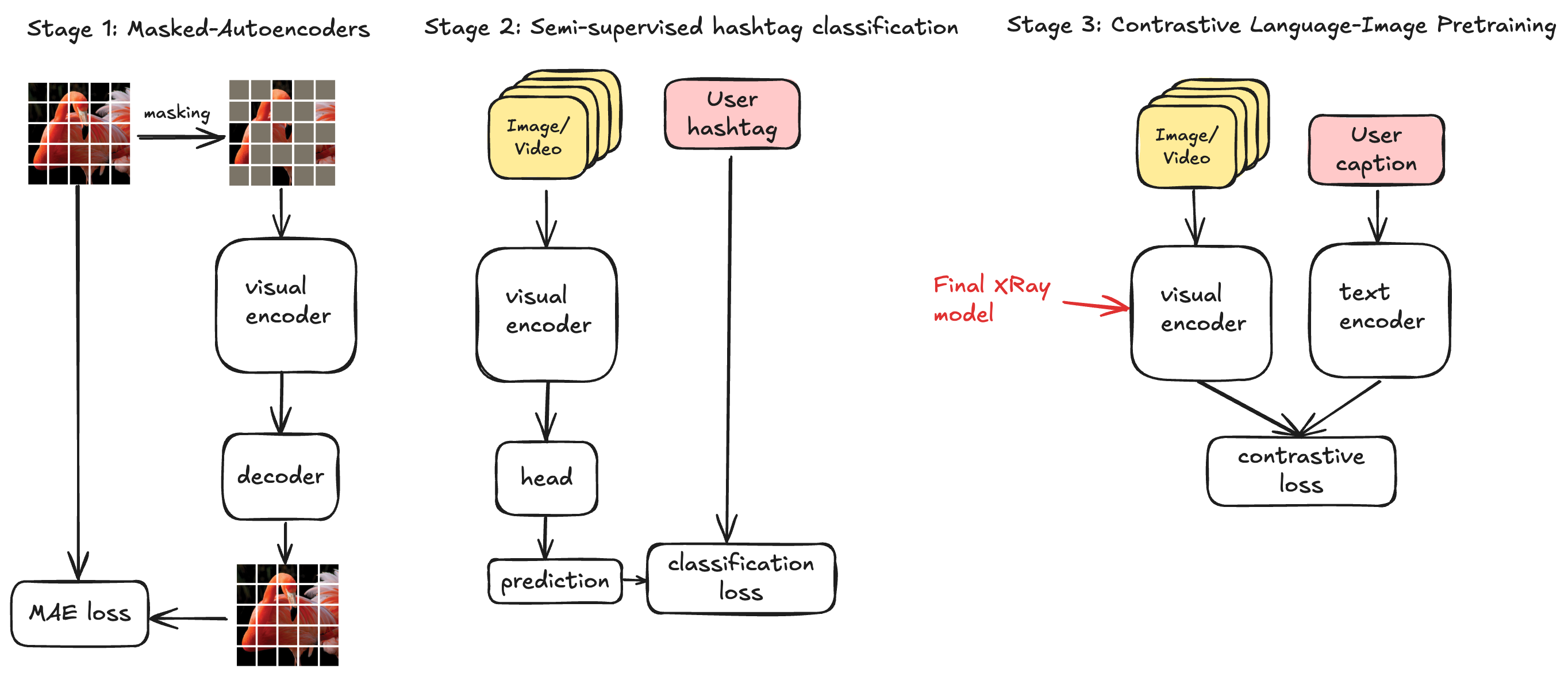}
    \caption{Our three-stage training pipeline consists of MAE pre-training, followed by hashtag classification, and followed by CLIP-style contrastive learning.}
    \label{fig:training_stages}
\end{figure}

Our training strategy consists of a three-stage pre-training process: (1) self-supervised learning via Masked Autoencoding (MAE)~\cite{mae}, (2) semi-supervised hashtag classification, and (3) semi-supervised contrastive learning with image/video-caption pairs. Figure~\ref{fig:training_stages} illustrates this progressive training pipeline.

The first two stages follow the MAE pre-pre-training framework described in~\cite{singh2023effectiveness}. Prior work has demonstrated that MAE pre-training exhibits favorable scaling properties with respect to both dataset and model size, and that combining MAE with weak supervision further enhances large-scale vision models~\cite{singh2023effectiveness}. This combination not only accelerates convergence but also provides a simple and scalable approach for learning visual representations at scale.

\subsubsection{Stage 1: Masked Autoencoding}

MAE~\cite{mae} learns visual representations from unlabeled image and video datasets through reconstruction. We adopt MAE as our initial stage due to its strong scaling behavior with respect to both model and dataset size. The MAE framework randomly masks 75\% of image patches and trains the model to reconstruct the masked input by minimizing pixel reconstruction error. For videos, we apply a higher masking ratio of 90\% following \cite{feichtenhofer2022maskedautoencodersspatiotemporallearners}. The target pixel values for each patch are normalized by the patch-wise mean and standard deviation.

Leveraging the ViT architecture, MAE training operates on only the 25\% unmasked patches, while a separate lightweight decoder reconstructs the missing regions. This asymmetric encoder-decoder design enables highly efficient training and supports the use of larger visual encoders.

\subsubsection{Stage 2: Semi-Supervised Hashtag Classification}

Following MAE pre-training, we employ hashtag classification. We find this semi-supervised approach highly effective, yielding results comparable to training on dedicated video and image baselines respectively. The data for both the first and second stages is sourced from the URU nouns hashtags dataset, enabling the model to learn robust object-centric representations. We can see in Table \ref{tab:training_ablation} row-1 that our hashtag finetuned model is pretty strong and gives very sold Kinetics and ImageNet performance.

\subsubsection{Stage 3: Contrastive Learning with Captions}

In the final stage, we introduce CLIP-style training~\cite{clip} using image and video caption data. CLIP is well-established for its strong generalization capabilities in computer vision and exhibits excellent scaling behavior, which we corroborate in our internal experiments. For this stage, we leverage ViSE data curated from user captions for images, and URU actions data for videos, as the latter emphasizes action understanding. Notably, the data used in the first two stages primarily focuses on noun-centric visual concepts, while the third stage emphasizes verb-centric action understanding, enabling the model to learn comprehensive representations encompassing both objects and actions. We can see in Table \ref{tab:training_ablation} row-2 has very strong zero shot generalization due to clip training and strong linear probing.

Table~\ref{tab:training_ablation} presents ablation results for different pre-training strategies using the ViT-H backbone, demonstrating the progressive improvements from each stage and additional techniques.

\begin{table}[h!]
\centering
\begin{tabular}{|l|c|c|c|}
\hline
\textbf{Training Configuration} & \makecell{\textbf{Spatial} \\ \textbf{Size}} & \makecell{\textbf{K700} \\ \textbf{Top-1}}  & \makecell{\textbf{IN} \\ \textbf{Top-1}} \\
\hline
MAE $\rightarrow$ URU & 280 & 75.1 & 87.1  \\
\hline
MAE $\rightarrow$ URU $\rightarrow$ CLIP & 280 & 75.2  & 87.3  \\
\hline
+ Registers & 280 & 75.4  & 87.3 \\
\hline
+ Augmentations & 280 & 76.1  & 87.6  \\
\hline
+ SLIP & 280 & 76.7  & 87.8 \\
\hline
\end{tabular}
\caption{Ablation study on training stages and techniques using ViT-H. K700: Kinetics-700, IN: ImageNet.}
\label{tab:training_ablation}
\end{table}

\subsubsection{SLIP: Combining Self-Supervised and Supervised Learning}

In addition to CLIP, we experiment with SLIP~\cite{SLIP}, which integrates self-supervised and contrastive learning objectives. We utilize SimCLR~\cite{simclr}, a self-supervised contrastive learning method that has demonstrated significant progress in computer vision, often matching or surpassing supervised learning in tasks such as image recognition, object detection, and semantic segmentation~\cite{dino,simclr}. Self-supervised image features have proven highly effective for grounding vision-language models (VLMs) and multimodal large language models (MM-LLMs)~\cite{cambrian}, and are widely adopted for their generalization capabilities and robustness.

We employ SimCLR as our primary self-supervised learning module due to its demonstrated scalability on large-scale datasets such as JFT~\cite{Mishra2022ASE}. SimCLR applies stochastic data augmentations—including random cropping, color distortions, and Gaussian blur—to generate pairs of correlated views from the same image, which serve as positive pairs. These augmented images are processed through a base encoder (typically a CNN or ViT), followed by a projection head that maps representations to a lower-dimensional space where contrastive loss is applied. The projection head, implemented as a small multi-layer perceptron, facilitates the learning of discriminative features.

The contrastive loss maximizes agreement between embeddings of positive pairs while minimizing similarity to negative pairs (embeddings from different images). This objective encourages the model to learn invariant and robust features without relying on labeled data, significantly improving representation quality.

While standard SimCLR uses two random crops of the same image as positive samples, we extend this approach for video data to better capture temporal information: we sample two frames from the same video and treat them as positive pairs, enabling the model to learn temporally invariant video representations.


\subsubsection{Denoising loss}
In addition to the CLIP loss, we incorporate a denoising loss $\mathcal{L}_{den}$ into our training objective. The denoising loss serves two primary purposes: (1) it stabilizes the training process and encourages the model to learn more robust visual representations and (2) it enhances scalability with respect to the number of training examples. This approach is motivated by prior work, which demonstrates the effectiveness of denoising objectives in large-scale vision models \cite{chen2024deconstructing}. While \cite{chen2024deconstructing} focuses on deconstructing diffusion models into denoising networks for self-supervised learning on image data, our approach expands the applicability of the denoising loss in two significant ways. First, we directly inject noise to the XrayVisual output. This is unlike \cite{chen2024deconstructing}, which adds noise to the PCA embedding space. This is crucial because the embedding space of the visual encoder is continually updated during training, which in turn modifies the eigenvector space. As the objective of the denoising encoder-decoder keeps getting moved with every iteration, it results in a less than optimal optimization. We validate this with subpar ImageNet accuracies on an ablation with a PCA bottleneck. By avoiding PCA, we ensure that the denoising process remains consistent and robust throughout training, regardless of changes in the encoder’s representation space.
Second, we create an effective information bottleneck first down-projecting the XrayVisual embedding to 0.1x the size using an MLP encoder. We then add zero-mean random gaussian noise to the down-projected embedding followed by up-projecting MLPs as a decoder to reconstruct denoised embeddings. We use L2 norm as the reconstruction loss function which we denote as $\mathcal{L}_{den}$. This architecture allows us to learn a more flexible and expressive mapping for denoising, tailored to the evolving feature space of the visual encoder.

\noindent \textbf{Robust Representations:} Models trained with the denoising loss exhibit greater robustness to noise and perturbations in the input data. On a training run with 600k iterations, we observe that the ImageNet accuracy increases by 0.3\% from 79.09\% to 79.39\% when the denoising loss is included. 

\noindent \textbf{Scalability:} By adding the denoising loss, we also observe improved scalability as the number of training examples increase. This is particularly important for large-scale datasets, where contrastive losses may struggle to fully utilize the available data. 

Our scaling experiments address the following research questions:

\paragraph{Scaling with Training Examples:} Does the denoising loss improve model performance as the number of training examples increase? 

\begin{figure}[ht]
    \centering
    \includegraphics[width=0.5\textwidth]{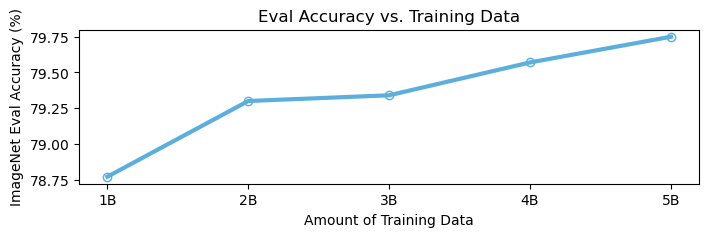}
    \caption{\textbf{ Scaling analysis.} ImageNet zero-shot accuracy improves consistently as the number of unique training examples increases from 1B to 5B. EViT-2B models trained with denoising loss exhibit strong scaling behavior across all data regimes. All models are trained for 500K iterations.}
    \label{fig:denoising:scaling}
\end{figure}

The impact of the denoising loss is illustrated in Figure \ref{fig:denoising:scaling} which show overall improvement in ImageNet Accuracies as the number of unique examples in training data increase of 1B through 5B. Note that we run all models for the same number of iterations. Hence, in ablations with fewer training examples than number of iterations, we cycle through the same data again till we reach a total of 500k iterations.

\paragraph{Scaling with Model Size:} Does the denoising loss have a similar scaling effect when applied to models of different sizes?

\paragraph{Importance of the Denoising Loss Scaling Parameter}
The scaling parameter for the denoising loss, denoted as $\lambda_{den}$
is a critical hyperparameter in our setup. Through empirical analysis, we find that setting $\lambda_{den}=1$ yields the best performance, providing a 0.3\% accuracy gain over the XrayVisual model trained solely with the CLIP loss.
\subsection{Progressive resolution} Following the successful paradigm established by SiGLIP~\cite{siglip, Zhai2023SigmoidLF} and Perception Encoder~\cite{bolya2025PerceptionEncoder}, we implement progressive resolution training during our CLIP model development. This multi-stage training approach enables efficient scaling while maintaining computational tractability across different resolution regimes.
Our training curriculum consists of three distinct phases: the initial and secondary training stages operate at 224×224 resolution to establish foundational representations. The final training stage employs a progressive resolution schedule, systematically increasing input resolution through the sequence: 98×98 → 154×154 → 224×224 → 336×336 → 448×448. This graduated approach allows the model to adapt incrementally to higher-resolution inputs while leveraging previously learned features.
We conducted comparative analysis with the FLIP (Fast Language-Image Pre-training)~\cite{Li2022ScalingLP} methodology to evaluate alternative scaling strategies. Our empirical evaluation demonstrates that progressive resolution training achieves superior computational efficiency compared to the FLIP approach, while maintaining competitive performance across evaluation benchmarks. The progressive resolution strategy proves particularly advantageous for large-scale training scenarios, offering an optimal balance between computational cost and model capability development. 
\subsection{Registers}
In the development of CLIP \cite{clip} and DINOv2 \cite{dino}, it was observed that certain tokens, referred to as outlier tokens, exhibit a norm approximately 10 times higher at the output and constitute a small fraction of the total sequence (around 2\%). These tokens tend to emerge around the middle layers of the vision transformer and only become apparent after extensive training with sufficiently large transformers. Since XrayVisual involves large-scale CLIP training, we observed similar token behavior in XrayVisual. The introduction of registers in DINOv2, as detailed in \citep{Registers, Oquab2023DINOv2LR}, presents a significant advancement in handling image recognition tasks within large-scale training pipelines like XrayVisual. Registers are additional tokens appended to the token sequence of a vision transformer, independent of the input image. This modification addresses specific challenges observed in the DINOv2 model, particularly concerning the presence of these outlier tokens. After adding registers into our pipeline, we saw improvements in video understanding. We measured XrayVisual's performance by evaluating image classification on ImageNet for image understanding and Kinetics-700 for video temporal understanding. On Kinetics-700, we observed a gain of 0.8\% and a slight improvement (+0.3\%) in image accuracy.

\subsection{Lion optimizer}
We adopt Lion~\cite{Chen2023SymbolicDO} as our optimizer, which was discovered through algorithmic program search. Prior work has demonstrated that Lion achieves a 2\% improvement in ImageNet zero-shot accuracy for the BASIC model compared to Adafactor~\cite{Shazeer2018AdafactorAL}. To validate its effectiveness for our architecture, we conduct a direct comparison between Lion and AdamW optimizers.
Following the recommendations in~\cite{Chen2023SymbolicDO}, we configure Lion with momentum parameters $\beta_1 = 0.9$ and $\beta_2 = 0.99$. With this configuration, we observe a 0.3\% improvement in ImageNet zero-shot accuracy when using Lion compared to AdamW, confirming the optimizer's effectiveness for large-scale vision-language pre-training.

\subsection{LLM as text encoder}
 We were currently using standard CLIP text encoder used for contrastive training. However it has has a couple limitations:
Limited context window: can only process 77 tokens.
Weak text comprehension: the text encoder is small and may not be able to understand the text describing the image/video.
To solve this we initially tried scaling the text encoder and the finally we decided to use LLaMA-1b as our text encoder.

\subsubsection{Scaling text encoder}
The table below shows the improvement on ImageNet using larger text encoder in CLIP framework. We followed the setting presented in GPT-2 to increase the attention heads and transformer blocks. CLIP used a 12-layer-8-head base model as the text encoder. With increasing size, the model achieves better image-text alignment and hence higher zero-shot (ZS) accuracy while barely improves image encoder with marginal improvement on linear probing (LC) accuracy.

\begin{table}[h!]
\centering
\begin{tabular}{|c|c|c|c|c|}
\hline
\textbf{Heads} & \textbf{Layers} & \textbf{Embedding Dimension} & \textbf{LC (\%)} & \textbf{ZS (\%)} \\
\hline
8  & 12 & 512  & 83.05 & 71.24 \\
12 & 12 & 768  & 83.11 & 71.71 \\
16 & 24 & 1024 & 83.11 & 71.91 \\
20 & 36 & 1280 & 83.08 & 72.13 \\
\hline
\end{tabular}
\caption{Model Configuration and Performance}
\end{table}
However we still didn't see improved performance with scaling text encoder; hence we started moving to LLM based CLIP models which we describe next.
\subsubsection{LLaMA-1b text encoder}
Large language models (LLMs)~\cite{llama, gpt, gpt3} demonstrate exceptional capabilities in processing extensive text corpora and handling long, complex sequences. Recent work in LLM2CLIP~\cite{Huang2024LLM2CLIPPL} has shown that leveraging LLMs as text encoders yields significant improvements in retrieval performance. Building upon this foundation, we integrate LLMs as text encoders in our XRayVisual CLIP training to enhance retrieval capabilities. This approach offers additional advantages including support for long, dense captions and inherent multilingual capabilities.
\paragraph{LLM2CLIP Methodology.} The LLM2CLIP framework consists of two primary components. Initially, the LLM undergoes fine-tuning using LoRA~\cite{Hu2021LoRALA} to achieve proper output space alignment. A critical observation from LLM2CLIP research reveals that vanilla LLM output features for different texts exhibit high similarity, resulting in insufficient discriminative power. This feature homogeneity severely impairs CLIP's contrastive learning objective, leading to suboptimal vision encoder performance when employing off-the-shelf LLMs.
\paragraph{Alignment and Fine-tuning Strategy.} To address the discriminative power limitation, we adopt the LLM2Vec approach, which expands the LLM's attention mechanism to bidirectional processing and employs masked next token prediction for improved output space alignment of semantically similar texts. Subsequently, we perform caption contrastive fine-tuning using supervised SimCSE~\cite{Gao2021SimCSESC} on MS-COCO~\cite{coco} captions and re-annotated captions generated by ShareCaptioner. This process optimizes the embedding space by bringing captions of identical images closer together while separating captions from different images.
\paragraph{Implementation Details.} Our LLM2CLIP implementation utilizes LLaMA-1B~\cite{llama, Touvron2023LLaMAOA} as the text encoder backbone. During training, we fine-tune the CLIP vision encoder while maintaining the fine-tuned LLM in a frozen state. This freezing strategy provides dual benefits: reduced memory consumption critical for maintaining large batch sizes necessary for effective negative sampling, and preservation of the LLM's comprehensive open-world knowledge.
To enhance alignment between the LLM and CLIP visual encoder, we incorporate an adapter module serving as learnable parameters. Following~\cite{Huang2024LLM2CLIPPL}, we implement a 4-layer adapter architecture. Projection layers are subsequently employed to ensure dimensional compatibility between the two encoder modalities.
Our training regimen incorporates both image-text and video-text pairs, with comprehensive experimentation across synthetic and user-generated caption datasets.

\begin{figure}[ht]
    \centering
    \includegraphics[width=0.5\textwidth]{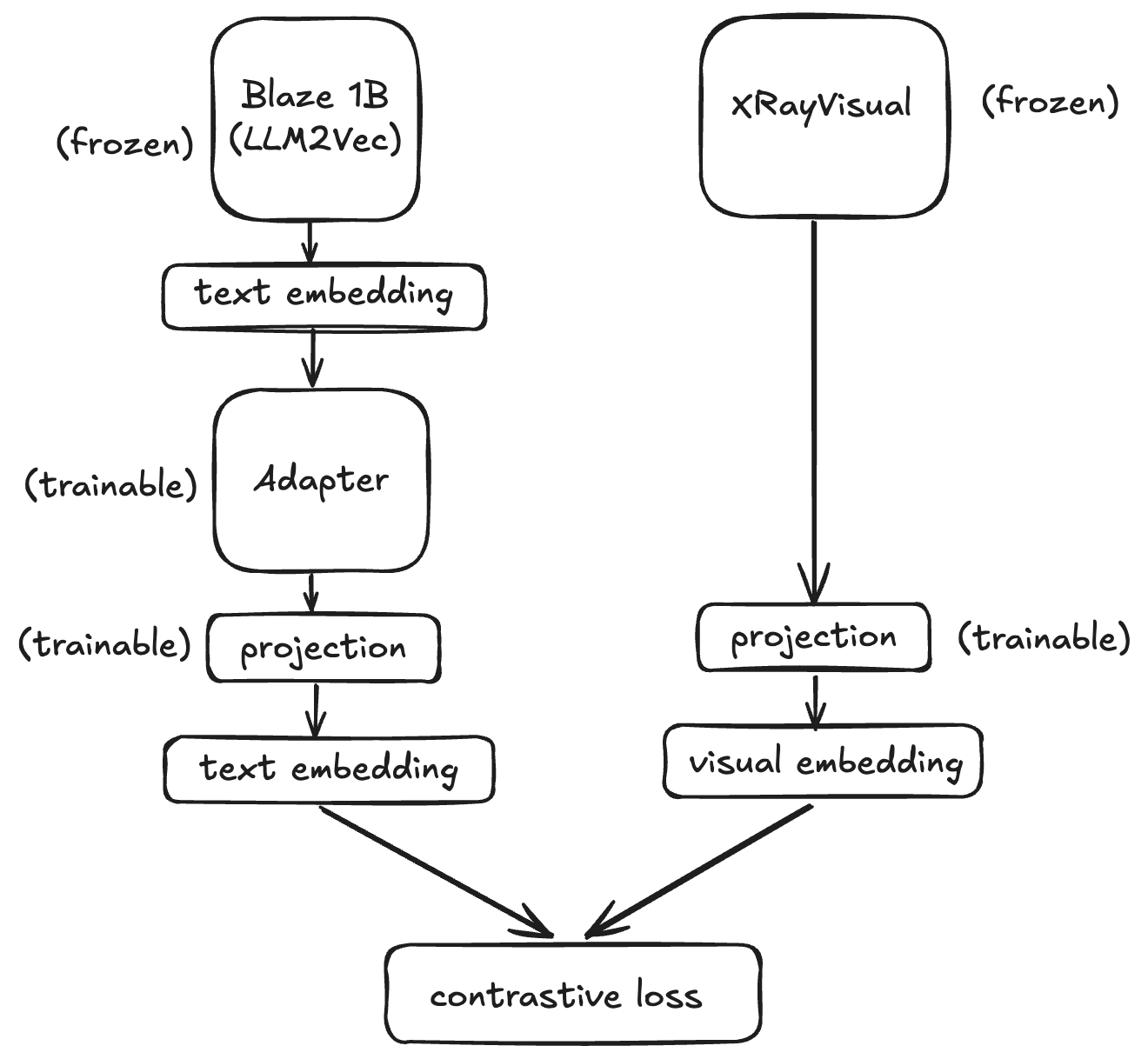}
    \caption{llm2clip }
    \label{fig:your_label}
\end{figure}

\begin{table}[!ht]
\centering
\caption{Using LLM as text encoder zero-shot performance. Both of these models use ViT-H for vision encoder.}
\label{tab:model_comparison}
\resizebox{\textwidth}{!}{%
\begin{tabular}{|l|c|c|c|c|c|c|c|c|c|}
\hline
\textbf{Models} & \makecell{}\textbf{COCO Img-Text r@1} & \textbf{COCO Text-Img r@1} & \textbf{MSRVTT Vid-Text r@1} & \textbf{MSRVTT Text-Vid r@1} & \textbf{ImageNet top-1} & \textbf{K700 top-1} & \textbf{Ads2Ads} & \textbf{IG Reels2Ads} & \textbf{FB Reels2Ads} \\
\hline
XRay & 48.21 & 31.12 & 51.4 & 27.92 & 79.19 & 61.53 & 88.27 & 62.42 & 70.76 \\
\hline
LLM2CLIP & 49.71 & 37.1 & 58.2 & 31.9 & 77.05 & 60.52 & 89.76 & 67.04 & 74.96 \\
\hline
\end{tabular}
}
\end{table}

\subsection{Efficiency with Token Merging}
One of the primary objectives for deploying models at large scale is efficiency. To this end, we employ token merging to reduce computational overhead.
We utilize ToMe (Token Merging)~\cite{Bolya2022TokenMY} to minimize the number of tokens processed in various layers of the ViT architecture, thereby reducing computation. This approach can be applied during both training and inference. Notably, we find that applying ToMe exclusively during inference results in minimal performance regression.
Specifically, using ToMe with ViT-H at a spatial resolution of 280 during inference leads to less than a 0.5\% drop in ImageNet accuracy and only a 0.04\% drop on Kinetics, while video inference speed increases by 58\%.

\subsubsection{Synthetic Caption Generation and Refinement with Siamese loss}

We utilize an internal multimodal large language model (MMLLM) to generate synthetic captions for videos. Despite the inherent noise in automatically generated captions, we observe substantial improvements, achieving a 6\% gain on MSR-VTT media-to-text retrieval compared to using user-generated captions alone.

\paragraph{LLM-Based Caption Refinement}
Initial analysis revealed that a small fraction (less than 1\%) of MMLLM-generated captions exhibited hallucinations, repetitive phrases, or excessive length. To address these issues, we employ LLaMA for chain-of-thought rewriting, which proves particularly effective at summarizing verbose captions and eliminating repetitive content. This refinement yields notable improvements: a 4\% boost in MSR-VTT media-to-text retrieval and a 0.2\% increase on ImageNet zero-shot accuracy.

\paragraph{Addressing Action Understanding Trade-offs}
However, we observe a 4\% performance drop on the Kinetics dataset when training exclusively with synthetic captions. We attribute this degradation to the loss of hashtag information, which is crucial for action understanding but often not captured in synthetic captions. To recover this performance while maintaining the benefits of synthetic captions, we introduce a siamese loss architecture operating on dual caption types:

\begin{itemize}
    \item \textbf{Synthetic caption loss}: Computed on MMLLM-generated captions, where we occasionally sample a single sentence instead of the full caption (following the approach in~\cite{bolya2025PerceptionEncoder}).
    \item \textbf{Hashtag caption loss}: Computed on aggregated hashtags from the original posts, preserving action-centric semantic information.
\end{itemize}

The siamese loss formulation allows the model to jointly learn from both caption types, successfully recovering the Kinetics performance while maintaining improvements on retrieval tasks. Through systematic experimentation with different loss weighting schemes, we find that a 1:1 ratio between synthetic caption loss and hashtag caption loss yields optimal results across all benchmarks.
\begin{figure}[ht]
    \centering
    \includegraphics[width=0.75\columnwidth]{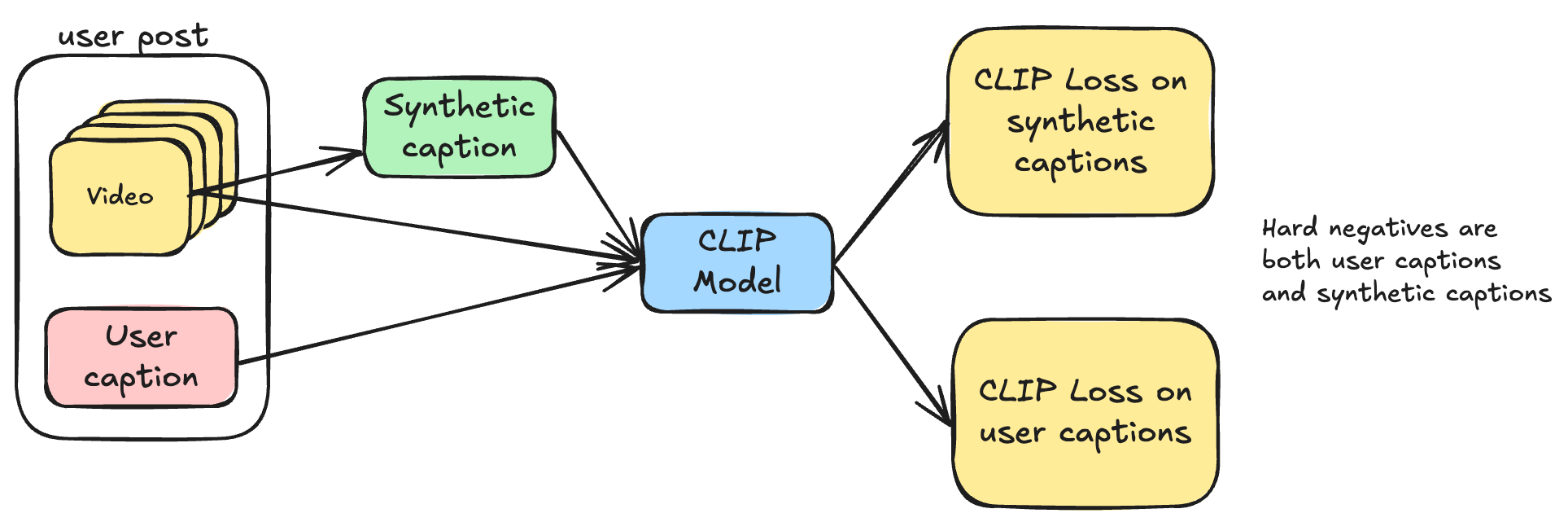}
    \caption{Siamese Loss}
    \label{fig:siamese_loss}
\end{figure}
 
\subsection{Attention pooling} We also tried attention pooling \cite{poolingattentioneffectivedesigns} in Xray , however we didn’t see any significant gains by using  attention pooling. This is most likely due to the scale of the data that has been used, which is 15B samples i.e. 8X as compared to PE \cite{bolya2025PerceptionEncoder}, that we don't see any upsides. Similarly we tried RoPE \cite{Su2021RoFormerET} poistional embedding as well, but that also didn't result into any significant gains.

\subsection{Data augmentations} We implement a comprehensive data augmentation pipeline that builds upon established techniques while incorporating domain-specific enhancements. Our augmentation strategy adopts proven methods from Perception Encoder (PE)~\cite{bolya2025PerceptionEncoder}, including brightness and saturation jitter transformations, as well as horizontal flipping operations.
Beyond these standard augmentations, we introduce Gaussian blurring as an additional preprocessing. This augmentation addresses a key challenge in real-world deployment scenarios where input imagery frequently exhibits substantial blurring or compression artifacts due to varying capture conditions.
The complete augmentation pipeline is integrated with our self-supervised learning objective, creating a robust training regime that enhances model generalization capabilities. This comprehensive approach yields measurable performance improvements, contributing to a 0.3\% accuracy gain on the Kinetics dataset. The incorporation of blur-specific augmentation proves particularly valuable for systems like Xray, where maintaining performance across diverse input quality conditions.

\subsection{Native resolution Support with Variable aspect ratio}
 Contemporary vision encoders~\cite{siglip, Zhu2025InternVL3EA, bolya2025PerceptionEncoder} typically process full-resolution images and videos to achieve optimal performance, avoiding the information loss that results from cropping—a well-documented limitation in vision models. To address this challenge, we introduce a Variable Aspect Ratio Image Transform that preserves critical visual information while maintaining computational efficiency.
Our proposed transform operates by intelligently resizing images along their larger dimension to conform to a specified target resolution while preserving the original aspect ratio. The method computes optimal scaling factors for both width and height dimensions, ensuring proportional resizing that maintains the image's geometric properties. Specifically, the transform resizes the input such that either the width or height completely occupies the target dimension, determined by whichever dimension is larger in the original image.
To achieve the desired output resolution, the transform applies padding with a black background to fill the remaining spatial dimensions, with the resized image positioned at the center of the output frame. This approach effectively eliminates information loss that would otherwise occur through aggressive cropping or aspect ratio distortion, thereby preserving the visual integrity and semantic content of the input imagery.
The Variable Aspect Ratio Image Transform thus enables our model to process diverse image formats without compromising on visual fidelity or computational performance, contributing to enhanced overall model effectiveness.


\subsection{Knowledge Distillation}

We employ a two-stage distillation strategy, first training large-scale 2B parameter EViT models as teachers, then distilling their knowledge into more efficient ViT-L student models. This approach enables us to achieve strong performance while maintaining computational efficiency for deployment.

\paragraph{Teacher Model Training}
We train separate teacher models for image and video modalities, each leveraging the strengths of our multi-modal training pipeline:

\noindent \textbf{Image Teacher:} We jointly train on the MetaCLIP and ViSE datasets to create a robust image teacher model. This combination leverages the diversity of MetaCLIP's web-scale data with the high-quality social media captions from ViSE.

\noindent \textbf{Video Teacher:} For the video teacher, we combine ViSE image data with our video datasets using a 75:25 ratio. This mixed-modality training approach proves more effective than video-only training, as the model benefits from the rich visual concepts learned from images while developing temporal understanding from videos.

\paragraph{Distillation Strategy}
We distill both teacher models into ViT-L architectures for practical deployment. While we experimented with video-only teacher models, we observe superior results when the teacher is trained on a combined image-video recipe, validating our multi-modal training approach. The knowledge from large-scale joint training effectively transfers to smaller models, enabling efficient inference while preserving the representational quality learned at scale.

\section{Results}
We present comprehensive experimental results demonstrating the effectiveness of our approach trained on large-scale social media datasets. Our method achieves state-of-the-art performance across multiple challenging benchmarks, spanning image classification, video understanding, and cross-modal retrieval tasks. Our model establishes new state-of-the-art results on ImageNet \cite{imagenet} classification, demonstrating superior performance compared to existing approaches. Furthermore, we evaluate robustness across challenging distribution shift scenarios, achieving state-of-the-art performance on established robustness benchmarks including ObjectNet \cite{Barbu2019ObjectNetAL} and ImageNet-Sketch \cite{wang2019learning}, which test model generalization under domain shift and stylistic variations. Extending our approach to video understanding through CLIP-style training methodologies, we achieve state-of-the-art results on standard video classification benchmarks, including Kinetics \cite{Carreira2017QuoVA} and HMDB51 \cite{Kuehne2011HMDBAL}. These results demonstrate the effectiveness of our training paradigm for temporal reasoning and action recognition tasks. Our method demonstrates exceptional performance on cross-modal retrieval tasks, establishing state-of-the-art results on both MSRVTT \cite{Xu2016MSRVTTAL} and MSCOCO \cite{coco} retrieval benchmarks. These results validate the quality of our learned joint embedding space for vision-language understanding. We conduct extensive evaluation on internal benchmarks to assess real-world applicability. Importantly, our analysis reveals a critical finding that challenges conventional assumptions: achieving state-of-the-art performance on established academic benchmarks does not necessarily translate to consistent improvements on internal evaluation metrics. This observation highlights the importance of comprehensive evaluation beyond standard benchmarks for practical deployment scenarios.
\subsection{SOTA results} We now present our SOTA results across linear probing, zero shot and then retrieval on internal metrics.
\subsubsection{Linear probing} 
\paragraph{Linear Probing Performance.} Table~\ref{ref:linear_probing_main} presents our comprehensive linear probing evaluation results. Our best-performing model, trained on the combination of ViSE and MetaCLIP datasets, achieves 89.3\% accuracy on ImageNet linear probing, establishing a new state-of-the-art for this benchmark. Notably, our model trained exclusively on ViSE data without any external dataset augmentation achieves 89.1\% accuracy, demonstrating the high quality and sufficiency of our curated training data.

\begin{table}[h!]
\centering
\footnotesize
\renewcommand{\arraystretch}{1.1}
\begin{tabularx}{\textwidth}{|X|X|c|c|c|c|}
\hline
\textbf{Name} & \textbf{Encoder} &  \textbf{ImageNet Top-1} & \textbf{ImageNet Top-5} & \textbf{Kinetics Top-1} & \textbf{Kinetics Top-5}  \\
\hline
PE-L      & PE-L    & 87.64 & 98.53 & 72.1 & 90.32 \\
\hline
PE-G      & PE-G    & 89.22 & 98.53 & 76.68 & 92.79 \\
\hline
SiGLIP-L  &   ViTL-16    & 83.10  & - & - & -     \\
\hline

DiNO-v3 & 7B   & 88.10  & -  & 68.10 &  84.51    \\

\hline
X-Ray ViTL-16   & ViTL-16 & 87.94 & 98.51 & -   & -   \\

\hline
X-Ray Image EViT-2b   & EViT-2b & \textbf{89.30} & \textbf{97.37} & -  & -   \\
\hline
X-Ray Visual EViT-2b   & EViT-2b & \textbf{88.10} & \textbf{96.37} & \textbf{78.10}  & 98.37   \\
\hline
X-Ray Image (ViSE)   & EViT-2b & \textbf{89.19} & \textbf{97.14} & - & -   \\
\hline
\end{tabularx}
\caption{Model and ImageNet Linear probing performance metrics.}
\label{ref:linear_probing_main}
\end{table}

\begin{table}[h!]
\centering
\footnotesize
\renewcommand{\arraystretch}{1.1}
\begin{tabularx}{\textwidth}{|X|X|c|c|c|c|c|c|}
\hline
\textbf{Name} & \textbf{Encoder} &  \textbf{Kinetics Top-1} & \textbf{Kinetics Top-5} & \textbf{UCF101 Top-1} & \textbf{UCF101 Top-5}  & \textbf{HMDB51 Top-1} & \textbf{HMDB51 Top-5}  \\
\hline
PE-G      & PE-G    & 76.68 & 92.79 & 96.99 & \textbf{99.89} & \textbf{77.50} & \textbf{95.54} \\
\hline
SiGLIP 2  & ViT-g/16 & 51.15 & 69.96 & 55.31 & 68.93 & 47.83 & 73.87 \\
\hline
DiNO-v3 & ViT-H   & 70.19 & 88.36 & 94.30 & 99.52 & 65.58 & 91.75 \\
\hline
X-Ray Visual & EViT-2b & \textbf{78.10} & \textbf{98.37} & \textbf{98.24} & 99.86 & 74.69 & 94.21 \\
\hline
\end{tabularx}
\caption{Model and Video Classification Linear probing performance metrics.}
\label{ref:linear_probing_video}
\end{table}

\begin{table}[!ht]
\centering
\caption{Classification accuracy of linear probes trained on ImageNet1k with frozen backbones. Weakly- and self-supervised models are evaluated with image resolution adapted to 1024 patch tokens (i.e., $448 \times 448$ for patch size 14, $512 \times 512$ for patch size 16). For reference, we also list results from Dehghani et al. (2023) using a different evaluation protocol (marked with *). Xray is evauated on 336 resolution with only 288 active tokens; making it more efficient then the contemporary models, while getting improved results.}
\label{tab:classification_accuracy}
\begin{tabular}{lcccccccccc}
\toprule
& & \multicolumn{3}{c}{ImageNet} & \multicolumn{2}{c}{Rendition} & \multicolumn{3}{c}{Hard} \\
\cmidrule(lr){3-5} \cmidrule(lr){6-7} \cmidrule(lr){8-10}
Method & ViT & Val & V2 & ReaL & R & S & A & C $\downarrow$ & Obj. \\
\midrule
Supervised backbones & & & & & & & & & \\
Zhai et al. (2022a)* & G/14 & 89.0 & 81.3 & 90.6 & 91.7 & — & 78.8 & — & 69.6 \\
Chen et al. (2023)* & e/14 & 89.3 & 82.5 & 90.7 & 94.3 & — & 81.6 & — & 71.5 \\
Dehghani et al. (2023)* & 22B/14 & 89.5 & 83.2 & 90.9 & 94.3 & — & 83.8 & — & 74.3 \\
\midrule
Agglomerative backbones & & & & & & & & & \\
AM-RADIOv2.5 & g/14 & 88.0 & 80.2 & 90.3 & 83.8 & 67.1 & 81.3 & 27.1 & 68.4 \\
\midrule
Weakly-supervised backbones & & & & & & & & & \\
PEcore & G/14 & \textbf{89.3} & \textbf{81.6} & 90.4 & {92.2} & {71.9} & \textbf{89.0} & 22.7 & {80.2} \\
SigLIP 2 & g/16 & 89.1 & 81.6 & \textbf{90.5} & {92.2} & 71.8 & 84.6 & 30.0 & 78.6 \\
AIMv2 & 3B/14 & 87.9 & 79.5 & 89.7 & 82.3 & 67.1 & 74.5 & 29.5 & 69.0 \\
EVA-CLIP & 18B/14 & 87.9 & 79.3 & 89.5 & 85.2 & 64.0 & 81.6 & 33.0 & 71.9 \\
\midrule
Self-supervised backbones & & & & & & & & & \\
Web-DINO & 7B/14 & 85.9 & 77.1 & 88.6 & 75.6 & 64.0 & 71.6 & 31.2 & 69.7 \\
Franca & g/14 & 84.8 & 75.3 & 89.2 & 67.6 & 49.5 & 56.5 & 40.0 & 54.5 \\
DINOv2 & g/14 & 87.3 & 79.5 & 89.9 & 81.1 & 65.4 & 81.7 & 24.1 & 66.4 \\
DINOv3 & 7B/16 & 88.4 & 81.4 & 90.4 & 91.1 & 71.3 & 86.9 & 19.6 & 79.0 \\
\midrule

Ours & & & & & & & & & \\
Xray & EViT-2B & \textbf{89.3} & 81.4 & - & \textbf{94.4} & \textbf{72.9} &  \textbf{88.9} & \textbf{-} & \textbf{83.8} \\
\bottomrule
\end{tabular}
\end{table}

\paragraph{Computational Efficiency.} A key advantage of our approach lies in its computational efficiency. Our models demonstrate 4× improved efficiency compared to standard vision encoders, including Perception Encoder~\cite{bolya2025PerceptionEncoder}, CLIP~\cite{clip}, and DiNO~\cite{dino, Oquab2023DINOv2LR}, by utilizing only 25\% of the visual tokens during processing. This efficiency gain enables our method to achieve the dual objectives of scalability and state-of-the-art performance without compromising accuracy.
\paragraph{Video Classification Results.} For video understanding tasks, our approach achieves 78.1\% Top-1 accuracy on the Kinetics dataset, establishing state-of-the-art performance for video classification. Our method demonstrates substantial improvements over existing approaches, with gains of +1.5\% over Perception Encoder and +10\% over the DiNO vision encoder. We attribute these significant improvements on Kinetics to our comprehensive training methodology, which leverages over 5 billion video samples across a carefully designed three-stage training curriculum. This large-scale video training enables our model to capture rich temporal dynamics and complex motion patterns essential for robust video understanding.
\subsubsection{Zero Shot results} 
\paragraph{Zero-Shot Evaluation.} Table~\ref{ref:zero_shot_main} presents our comprehensive zero-shot evaluation results, demonstrating state-of-the-art performance across multiple benchmarks. Our method achieves superior zero-shot performance on both ImageNet Top-1 and MS-COCO Top-1 metrics, with notable improvements of +1.05\% on MS-COCO Top-1 and +0.25\% on ImageNet Top-1 compared to existing approaches.
To achieve these results, we employ a knowledge distillation strategy that transfers learned representations from our 2B parameter models to more efficient ViT-H and ViT-L architectures. This distillation process leverages synthetic captions generated using LLaMA-4 and ViSE-enhanced LLM-generated captions, which prove instrumental in improving zero-shot performance. While synthetic caption augmentation demonstrates clear benefits for zero-shot evaluation, we note that improvements in zero-shot metrics do not necessarily correlate with enhanced performance on internal retrieval benchmarks, as detailed in the subsequent section.
We also report results on new country211, fgvc-aircraft, food etc in Table \ref{tab:transfer_learning}. We see that our OOD generalization is pretty strong and we SOTA results on ImageNet-Sketch \cite{wang2019learning}, ImageNet rendidtion. Interestingly on some datasets like sun397 and resisc45, our results are not SOTA; this primarily is due to difference in data distibution between academic datasets and large scale industry datasets.    

\subsection{Transfer learning results} We also do transfer learning by evaluating our model on linear probing in transfer learning setting.




\begin{table}[h!]

\centering

\footnotesize

\renewcommand{\arraystretch}{1.1}

\begin{tabularx}{\textwidth}{|X|X|c|c|c|c|c|}

\hline

\textbf{Name} & \textbf{Encoder} & \textbf{Parameter Count}& \textbf{MSCOCO Img-Text r@1} & \textbf{MSCOCO Text-Img r@1} & \textbf{ImageNet Top-1} & \textbf{ImageNet Top-5} \\







\hline

PE      & ViTL & 300M &    72.75 & 54.21 & 83.08 & 97.22 \\

\hline

SiGLIP  &   ViTL & 300M &       71.40  & 55.30  & 83.10  & 97.20      \\

\hline



DiNO-v3.txt & ViT-7B & 7B    & 63.7  & 45.6  & 82.3  & 96.21      \\



\hline

X-Ray Visual   & ViTL & 300M  & 73.21 & 55.06 & 83.07 & 97.28 \\

\hline

X-Ray Visual   & ViTH & 832M & \textbf{73.83} & \textbf{55.11} & \textbf{83.33} & \textbf{97.37} \\

\hline

\end{tabularx}

\caption{Model and MSCOCO/ImageNet Zero-Shot performance metrics.}

\label{ref:zero_shot_main}

\end{table}

\subsubsection{MMEB Results} We evaluate our approach on the MMEB validation set through direct zero-shot inference without any fine-tuning on the MMEB training data. As demonstrated in Table~\ref{MMEB-ZS}, our Xray model achieves state-of-the-art results within the 2B parameter model category, often surpassing comparable methods by substantial margins. Specifically, we observe significant improvements of +8.1\% on ObjectNet, +15.5\% on ImageNet-Adversarial, and +4.1\% on ImageNet Rendition compared to the nearest competing approaches.
These substantial improvements on adversarial datasets highlight the robust generalization capabilities of our Xray model beyond its training distribution. This generalization behavior represents a critical advancement for production deployment scenarios involving large-scale Meta traffic, where models frequently encounter out-of-distribution (OOD) imagery. The strong performance on adversarial benchmarks validates the practical utility of our approach for real-world applications.

In Table \ref{tab:mmeb_video} we show results on video zers shot datasets. We get +5-6\% above the best multi-modal models and show promising gains while being 20x smaller then MMLLM's.

\begin{table}[!ht]
\centering

\resizebox{\textwidth}{!}{%
\begin{tabular}{|l|c|c|c|c|c|c|c|c|c|}
\hline
\textbf{Models} & \textbf{Model Size (B)} & \textbf{Image CLS} & \textbf{SUN397} & \textbf{ObjectNet} & \textbf{Country211} & \textbf{Place365} & \textbf{ImageNet-1K} & \textbf{ImageNet-A} & \textbf{ImageNet-R} \\
\hline
XRay Model EViT-2B & \textbf{2.4} & \textbf{66.8} & \textbf{71.9} & \textbf{79.4} & \textbf{44.2} & \textbf{44.8} & \textbf{79.5} & \textbf{68.6} & \textbf{95.3} \\
\hline
XRay Model ViT-H & \textbf{0.6B} & \textbf{66.1} & \textbf{73.5} & \textbf{79.6} & \textbf{36.0} & \textbf{45.7} & \textbf{83.3} & \textbf{65.2} & \textbf{93.6} \\
\hline
Ops-MM-embedding-v1-2B & 2.21 & 65.1 & 80.7 & 68.4 & 28.6 & 43.9 & 81.1 & 53.1 & 91.2 \\
\hline
RzenEmbed-v1-2B & 2.21 & 61.2 & 78.4 & 71.3 & 24.5 & 41.3 & 80.9 & 49.5 & 84.8 \\
\hline
VLM2Vec-V2.0-Qwen2VL-2B & 2.21 & 58.9 & 71.0 & 65.2 & 25.2 & 35.9 & 80.8 & 47.4 & 89.3 \\
\hline
VLM2Vec-V1-Qwen2VL-2B & 2.21 & 54.9 & 73.8 & 37.1 & 21.5 & 35.3 & 77.5 & 50.9 & 84.7 \\
\hline
gme-Qwen2-VL-2B-Instruct & 2.21 & 52.3 & 67.3 & 70.6 & 26.5 & 35.8 & 58.3 & 28.8 & 78.6 \\
\hline
interestFM-UIR-CAFe-0.5B & 0.894 & 51.6 & 68.8 & 51.0 & 11.3 & 37.4 & 64.6 & 38.1 & 86.4 \\
\hline
colpali-v1.3 & 2.92 & 38.5 & 56.1 & 45.6 & 6.0 & 27.5 & 42.4 & 14.9 & 64.6 \\
\hline

\end{tabular}
}

\caption{MMEB Performance Comparison of Models}

\label{MMEB-ZS}

\end{table}

\begin{table}[h!]
\centering
\resizebox{\textwidth}{!}{
\begin{tabular}{|l|c|c|c|c|c|c|c|c|c|c|c|c|c|c|}
\hline
 & cars & country211 & fgvc-aircraft & food101 & imagenet-a & imagenet-r & imagenet1k & imagenet-sketch & imagenetv2 & objectnet & sun397 & vtab-flowers & vtab-pets & vtab-resisc45 \\
\hline
KD 2B to ViT-H, Vise & 0.85040 & 0.34290 & 0.33710 & 0.96230 & 0.83980 & 0.93560 & 0.82200 & 0.85040 & 0.76440 & 0.76910 & 0.75120 & 0.89700 & 0.96810 & 0.63820 \\
\hline
KD 2B to ViT-H, M+V+L4 & 0.93710 & 0.39660 & 0.68150 & 0.96250 & 0.86260 & 0.94740 & 0.83330 & 0.93710 & 0.77970 & 0.81630 & 0.77350 & 0.88690 & 0.96400 & 0.71490 \\
\hline
MetaCLIP+ViSE & 0.90027 & 0.40137 & 0.41016 & 0.97012 & 0.86641 & 0.94458 & 0.82914 & 0.77764 & 0.72584 & 0.82221 & 0.74619 & 0.86507 & 0.97339 & 0.66376 \\
\hline
Only ViSE & 0.90027 & 0.40137 & 0.41016 & 0.97012 & 0.86641 & 0.94458 & 0.82914 & 0.77764 & 0.72584 & 0.82221 & 0.74619 & 0.86507 & 0.97339 & 0.66376 \\
\hline
LiT-22B & - & - & - & - & 0.901 & 0.960 & 0.859 & - & 0.809 & 0.876 & - & - & - & - \\
\hline
SigLIP-B/16 & 0.908 & 0.440 & 0.159 & 0.916 & 0.451 & 0.902 & 0.762 & 0.679 & 0.695 & 0.707 & - & 0.852 & 0.942 & 0.646 \\
\hline
SigLIP2-B/16 & 0.934 & 0.548 & 0.192 & 0.928 & 0.550 & 0.917 & 0.782 & 0.689 & 0.731 & 0.736 & - & 0.857 & 0.954 & 0.711 \\
\hline
PEcore B & 0.921 & 0.570 & 0.305 & 0.925 & 0.624 & 0.887 & 0.784 & 0.661 & 0.750 & 0.719 & - & 0.865 & 0.946 & 0.727 \\
\hline
SigLIP-L/16 & 0.948 & 0.532 & 0.247 & 0.956 & 0.765 & 0.950 & 0.736 & 0.744 & 0.956 & 0.894 & 0.968 & 0.948 & 0.532 & 0.679 \\
\hline
SigLIP2-L/16 & 0.958 & 0.670 & 0.316 & 0.961 & 0.843 & 0.957 & 0.755 & 0.784 & 0.961 & 0.900 & 0.964 & 0.958 & 0.670 & 0.755 \\
\hline
PEcore L & 0.937 & 0.678 & 0.456 & 0.962 & 0.890 & 0.952 & 0.734 & 0.800 & 0.962 & 0.872 & 0.964 & 0.937 & 0.678 & 0.757 \\
\hline
DFN-H+ & 0.960 & 0.725 & 0.379 & 0.962 & 0.796 & 0.936 & 0.733 & 0.805 & 0.962 & 0.916 & 0.968 & 0.960 & 0.725 & 0.759 \\
\hline
InternVL-C & 0.944 & 0.533 & 0.351 & 0.953 & 0.838 & 0.957 & 0.743 & 0.764 & 0.953 & 0.858 & 0.963 & 0.944 & 0.533 & 0.744 \\
\hline
EVA 18B & 0.949 & 0.597 & 0.431 & 0.958 & 0.873 & 0.957 & 0.747 & 0.788 & 0.958 & 0.860 & 0.961 & 0.949 & 0.597 & 0.769 \\
\hline
EVA 18B+ & - & - & - & - & 0.889 & 0.956 & 0.743 & - & - & - & - & - & - & - \\
\hline
SigLIP2-g-opt & 0.959 & 0.736 & 0.401 & 0.970 & 0.905 & 0.966 & 0.774 & 0.810 & 0.970 & 0.915 & 0.978 & 0.959 & 0.736 & 0.759 \\
\hline
PEcore G (image only) & 0.946 & 0.767 & 0.573 & 0.966 & 0.912 & 0.961 & 0.761 & 0.827 & 0.966 & 0.910 & 0.964 & 0.946 & 0.767 & 0.718 \\
\hline
PEcore G & 0.947 & 0.782 & 0.576 & 0.969 & 0.926 & 0.965 & 0.765 & 0.837 & 0.969 & 0.914 & 0.969 & 0.947 & 0.782 & 0.758 \\
\hline

\end{tabular}
}

\caption{Performance of models on Transfer learning.}
\label{tab:transfer_learning}
\end{table}

\begin{table}[h!]
\centering
\resizebox{\textwidth}{!}{
\begin{tabular}{|l|l|c|c|c|c|c|c|c|}
\hline
\makecell{\textbf{Models}} & 
\makecell{\textbf{Backbone}} & 
\makecell{\textbf{Overall}} & 
\makecell{\textbf{UCF101}} & 
\makecell{\textbf{Breakfast}} & 
\makecell{\textbf{K-700}} & 
\makecell{\textbf{HMDB51}} & 
\makecell{\textbf{sth-sth V2}} \\
\hline
Qwen2 VL & 7 Billion  & 57.6 & 78.6 & 37.200 & 55.6 & 63.9 & 53 \\
\hline
X-Ray visual & 400M  & \textbf{63.0} & \textbf{85.5} & \textbf{41.8} & \textbf{66.7} & 57.8 & \textbf{54.2} \\
\hline
\end{tabular}
}

\caption{Comparison of model performance across MMEB Video benchmarks.}
\label{tab:mmeb_video}
\end{table}

\subsection{Internal metrics}

\paragraph{Internal Evaluation on Benchmarks.} Beyond standard academic benchmarks, we conduct extensive evaluation on internal reels-to-ads and FB search retrieval tasks, which serve as critical indicators of real-world performance. These benchmarks provide essential insights into model behavior on practical scenarios.

\paragraph{Dataset and Evaluation Protocol.} Our ads evaluation encompasses ad-to-ad and Instagram/Facebook reels-to-ads retrieval datasets, designed to assess the model's ability to determine semantic similarity between advertisements and social media content. Ground-truth similarity labels are established through majority vote consensus from a minimum of three independent human annotators, ensuring reliable evaluation standards. We evaluate semantic similarity by computing cosine similarity between learned embeddings, providing a direct measure of representational quality. FB search retrieval evaluation measures the model's ability to retrieve the corresponding image or text. User click through rate is used to remove noisy image and text FB post pairs.

\paragraph{Performance Analysis.} Table~\ref{tab:reels2ads} and ~\ref{tab:fbsearch} presents our comprehensive results on these internal benchmarks. Compared to models trained exclusively on external datasets, XRay demonstrates substantial performance improvements across all evaluation scenarios.

\paragraph{Domain Shift Analysis.} A critical finding from our evaluation reveals that state-of-the-art vision encoders, including Perception Encoder (PE) and DiNO, exhibit significant performance degradation under domain shift conditions. While XRay achieves a +1.8 AUC improvement on same-domain ads-to-ads retrieval, the performance gap becomes dramatically more pronounced in cross-domain scenarios. Specifically, XRay achieves remarkable improvements of +10.8\% on Instagram reels-to-ads retrieval and +10.2\% on Facebook reels-to-ads retrieval compared to the Perception Encoder baseline. Although XRay ViT-L variant and PE-L have similar MSCOCO retrieval performance, XRay achieves +2\% gain on FB search metrics.

\paragraph{Implications for Real-World Deployment.} These results highlight a fundamental limitation in current state-of-the-art vision encoders: substantial performance degradation when confronted with real-world out-of-distribution (OOD) scenarios. This observation underscores the critical need for incorporating more diverse training data in vision encoder development. While training on curated, clean datasets may yield strong performance on academic benchmarks, such approaches fail to generalize effectively to practical use cases—a phenomenon well-documented in large language models but previously underexplored in vision encoders.

\paragraph{LLM-based Text Encoders.} We investigate the effectiveness of employing large language models as text encoders in our multimodal framework. Specifically, we replace conventional text encoders with LLaMA~\cite{llama}, leveraging its enhanced linguistic representation capabilities for improved cross-modal understanding.
Our experimental analysis reveals that LLM-based text encoding yields substantially stronger overall performance compared to standard text encoding approaches. While academic benchmark evaluations did not demonstrate significant improvements, we observed considerable gains on internal evaluation metrics. This discrepancy between academic and internal benchmark performance further reinforces the critical importance of employing generalized, robust models for practical deployment scenarios. The superior performance of LLM-based text encoding on internal metrics suggests that the richer linguistic representations learned by large-scale language models provide enhanced semantic alignment capabilities, particularly beneficial for real-world applications involving diverse and complex textual content.

We also evalued our model on CIFAR dataset which is a low resolution dataset. We found our results not to be SOTA, primarly due to mismatch of training domains and resolution. However that doesn't deter from getting good results on real world datasets.

\begin{table}[h!]
\centering
\footnotesize
\renewcommand{\arraystretch}{1.1}
\begin{tabularx}{\textwidth}{|X|X|c|c|c|}
\hline
\textbf{Name} & \textbf{Encoder} & \textbf{Ad to Ad AUC} & \textbf{IG Reels to Ads AUC} & \textbf{FB Reels to Ads AUC} \\
\hline
XRay-LLM2CLIP & ViTL-16 & \textbf{90.21} & \textbf{67.03} & \textbf{77.7} \\
\hline
PE-L   & ViTL-16 & 88.54 & 56.14 & 67.36 \\
\hline
PE-G   & ViT-G & 89.67 & 63.22 & 69.86 \\
\hline
DinoV2 with text  & ViTL-14 & 84.35 & 53.69 & 61.4 \\
\hline
\end{tabularx}
\caption{Reels and Ads performance metrics}
\label{tab:reels2ads}
\end{table}

\begin{table}[h!]
\centering
\footnotesize
\renewcommand{\arraystretch}{1.1}
\begin{tabularx}{\textwidth}{|X|X|c|c|c|c|}
\hline
\textbf{Name} & \textbf{Encoder} & \textbf{COCO Img-Text r@1} & \textbf{COCO Text-Img r@1} &\textbf{FB Img-Text r@1} & \textbf{FB Text-Img r@1} \\
\hline
XRay Model & ViTL-16 & 73.21 & 55.06 & \textbf{52.06} & \textbf{48.9} \\
\hline
PE-L & ViTL-16 & 75.9 & 57.1 & 49.39 & 47.02 \\
\hline
\end{tabularx}
\caption{FB Search retrieval metrics}
\label{tab:fbsearch}
\end{table}

\section{Practical Deployment and Applications}
This section examines the practical deployment considerations and real-world applications of our proposed model, highlighting integration strategies and performance trade-offs.
\subsection{Applications}
Our model serves as a component across multiple production systems, providing robust visual representations for diverse downstream tasks. The primary application domain centers on recommendation systems, where the core objective involves retrieving the most semantically relevant videos given input images or videos. In recommendation frameworks, the system leverages user interaction history to retrieve contextually appropriate content conditioned on previous engagement patterns.
Additional deployment scenarios include bidirectional ads-to-reels retrieval systems and comprehensive ads-to-ads matching pipelines. These applications demonstrate the versatility of our learned representations across diverse content modalities.
\subsection{Linear Adapter Integration}
A particularly effective deployment strategy involves training lightweight linear adapters on top of our pre-trained embeddings for specialized downstream tasks. This approach enables rapid task-specific adaptation while preserving the core representational capabilities. For instance, we demonstrate this methodology for AI-generated content detection, where we utilize LLaMA~\cite{llama} for data annotation and subsequently train linear classifiers for content authenticity assessment. This approach achieves performance comparable to multimodal large language models while requiring only a fraction of the computational resources.
\subsection{Dimensionality Reduction for Scalable Deployment}
The full-dimensional embeddings from ViT-H (1280d) and ViT-L (1024d) models present significant computational and storage challenges. To address scalability requirements, we implement 100-dimensional projection layers that substantially reduce costs while maintaining reasonable performance.
Empirical evaluation reveals that this dimensionality reduction incurs approximately 4\% performance degradation on ImageNet classification and a more substantial ~5\% drop on Kinetics video classification. Despite these performance trade-offs, the cost-to-performance ratio strongly favors the 100-dimensional embeddings.
\subsection{Quantization-Based Compression}
As an alternative to dimensionality reduction, we explore quantization techniques for embedding compression. Our quantization pipeline reduces embeddings to int8 precision, achieving superior performance compared to 100-dimensional projections while maintaining storage efficiency.
Quantized embeddings demonstrate notably smaller performance degradation, with approximately 3\% accuracy reduction on ImageNet compared to full-precision representations. However, practical deployment of quantized embeddings presents integration challenges, as downstream client systems require specialized support and special care needs to be taken to use quantized embdedding.

\subsection{Similarity focussed models}
We also build specliazed models for focussing on near deuplicate detection task. Our large scale we saw that current Xray models perfomed pretty well on near duplicate detection. However we were able to improve upon it by making a new model, where we started by pre-training on MAE \cite{mae} and then we followed by SimCLR \cite{simclr}. We saw combination of these models gave the best results for near duplicates. 

\subsection{Semantic ID}
Semantic IDs are discrete tokenizers derived from embeddings. They play a key role in recommendation systems, where MTML models are typically designed to consume both continuous and sparse ID features. Unlike traditional unique item IDs, semantic IDs enable models to generalize to unseen or similar items, promote balanced learning, and capture hierarchical content structures through a shared vocabulary of tokens generated from item content. We experimented with two variants of Xray-Visual tokenizers built upon VQVAE-V2 \citep{vqvae}.

The first variant adopts residual quantization \citep{rqvae}. Given an Xray embedding as input, the model encodes it and performs a $K$-level hierarchical vector quantization. At each level, the input residual vector is matched against a codebook by cosine similarity, producing a discrete codeword and a residual that is passed to the next level. The $K$ selected codeword embeddings are then summed to reconstruct the input. This hierarchical quantization scheme yields compact and expressive semantic IDs that capture both coarse- and fine-grained content structures.

The second variant employs product quantization \citep{pqvae}. Instead of applying quantization sequentially, the model splits the encoded Xray embedding into $K$ chunks, each of which undergoes vector quantization in parallel. The resulting $K$ codeword embeddings are concatenated to reconstruct the input. This parallel quantization scheme produces semantic IDs that represent distinct topics within the same input.

In practice, we found that combining the two semantic ID variants provides complementary benefits for downstream applications. To stabilize training and improve codebook utilization in both variants, we follow VQVAE-V2 \citep{vqvae} and update the codebook vectors using an exponential moving average (EMA) rather than gradient-based optimization. This EMA update mitigates codebook collapse and promotes smoother convergence. 

\section{Related Works}
\subsection{CLIP}
There has been many approaches proposed for pre-training the ViT for vision-language alignment. CLIP~\citep{clip,jia2021scaling} becomes the popular choice, due to its superior performance on multimodal understanding as studied in~\cite{cambrian}. Other methods leverage sigmoid loss like SigLIP~\citep{siglip,siglip2} or captioning loss like Cappa~\citep{tschannen2024image} are also popular; LocCa~\citep{wan2024locca} further incorporates bounding box coordinates. However, they need to train a full encoder-decoder transformer with smaller batch sizes, which is less efficient than CLIP. Also, the vision embeddings are not directly aligned with languages, so more limited for search or retrieval tasks. Therefore we focus on comparing against works on the CLIP approach. 

To enhance the fine-grained understanding of vision embeddings, some recent works~\citep{naeem2023silc,bica2024improving,dong2023maskclip} combine localization-enhancing unsupervised objectives with the CLIP loss. Alpha-CLIP~\citep{Sun_2024_CVPR} shows that the SAM~\citep{kirillov2023segment} can provide useful conditions for CLIP. Another  work, CLOC~\citep{chen2025contrastive} uses supervision on large-scale explicit pseudo-labeled detection data to train the CLIP encoder enable better local feature extraction. AIM-v2~\citep{fini2025multimodal} proposed to combine pixel-level reconstruction loss with the CLIP loss. Differently, we inject the localization knowledge via our proposed multi-stage training recipe with MAE to  without any external supervision.  

Another promising direction is enhancing the quality of the image captions through LLM re-writing~\citep{lai2024revisit,nguyen2023improving}, which has been proven improving the retrieval performance. We also incorporate high-quality synthetic captions on the native social media domains that not only enrich the visual information but also better capture users' interests and intention.

\subsection{Self-Supervised Learning}
Contrastive learning has emerged as a foundational approach in self-supervised computer vision, enabling models to learn general-purpose representations without the need for labeled data. Early works such as SimCLR \cite{simclr}, MoCo \cite{he2020momentum}, and DINO\cite{dino} demonstrated that contrastive objectives—by enforcing invariance to data augmentations and leveraging negative samples—can achieve state-of-the-art performance on image recognition tasks [\cite{oord2018representation, henaff2020data, he2020momentum, tian2020contrastive, chuang2020debiased, chen2021empirical}; . These methods were initially developed for ConvNet backbones, but subsequent research showed that, with proper architectural tuning, they could be successfully adapted to Vision Transformers (ViT) \cite{vit}. The key idea in contrastive learning is to spread the learned embeddings uniformly on the sphere, using negative samples to avoid trivial solutions \citealp{robinson2020contrastive, Mishra2022ASE, Ge2021RobustCL, wang2019learning} . 

\subsection{Masked Image Modeling and Autoencoding Methods}
The advent of Vision Transformers (ViT) \cite{vit}  has also spurred the development of masked image modeling techniques, which take inspiration from masked language modeling in NLP \citep{devlin2018bert, bert, llama} . In this paradigm, models learn to reconstruct masked portions of input images, thereby capturing rich spatial statistical dependencies. Notable methods such as MAE (Masked Autoencoder) \cite{mae}, BEiT \cite{bao2021beit} , and related approaches \cite{chen2022context, xie2022simmim} have shown that it is possible to pre-train ViTs efficiently by omitting masked tokens from the encoder, resulting in significant computational savings. Unlike contrastive learning, autoencoders encourage information preservation in the latent representations, which can lead to different feature learning dynamics. The efficiency and scalability of masked image modeling have made it a popular choice for large-scale vision pre-training, and our work similarly leverages these advances to build robust multimodal models.

\clearpage
\newpage
\bibliographystyle{assets/plainnat}
\bibliography{paper}

@String(CVPR= {IEEE Conf. Comput. Vis. Pattern Recog.})

@String(ICCV= {Int. Conf. Comput. Vis.})

@String(ECCV= {Eur. Conf. Comput. Vis.})

@String(ICLR = {Int. Conf. Learn. Represent.})

@String(CVPR  = {CVPR})

@String(ICCV  = {ICCV})

@String(ECCV  = {ECCV})

@String(ICLR  = {ICLR})

@article{vqvae,
  title={Generating diverse high-fidelity images with vq-vae-2},
  author={Razavi, Ali and Van den Oord, Aaron and Vinyals, Oriol},
  journal={Advances in neural information processing systems},
  volume={32},
  year={2019}
}

@inproceedings{rqvae,
  title={Autoregressive image generation using residual quantization},
  author={Lee, Doyup and Kim, Chiheon and Kim, Saehoon and Cho, Minsu and Han, Wook-Shin},
  booktitle={Proceedings of the IEEE/CVF conference on computer vision and pattern recognition},
  pages={11523--11532},
  year={2022}
}

@inproceedings{pqvae,
  title={Generating long semantic IDs in parallel for recommendation},
  author={Hou, Yupeng and Li, Jiacheng and Shin, Ashley and Jeon, Jinsung and Santhanam, Abhishek and Shao, Wei and Hassani, Kaveh and Yao, Ning and McAuley, Julian},
  booktitle={Proceedings of the 31st ACM SIGKDD Conference on Knowledge Discovery and Data Mining V. 2},
  pages={956--966},
  year={2025}
}

@inproceedings{gpt,
  title={Language models are few-shot learners},
  author={Brown, Tom and Mann, Benjamin and Ryder, Nick and Subbiah, Melanie and Kaplan, Jared D and Dhariwal, Prafulla and Neelakantan, Arvind and Shyam, Pranav and Sastry, Girish and Askell, Amanda and others},
  booktitle={NeurIPS},
  year={2020}
}

@inproceedings{clip,
  title={Learning transferable visual models from natural language supervision},
  author={Radford, Alec and Kim, Jong Wook and Hallacy, Chris and Ramesh, Aditya and Goh, Gabriel and Agarwal, Sandhini and Sastry, Girish and Askell, Amanda and Mishkin, Pamela and Clark, Jack and others},
  booktitle={ICML},
  year={2021},
}

@article{gemma,
  title={Gemma: Open models based on gemini research and technology},
  author={Team, Gemma and Mesnard, Thomas and Hardin, Cassidy and Dadashi, Robert and Bhupatiraju, Surya and Pathak, Shreya and Sifre, Laurent and Rivi{\`e}re, Morgane and Kale, Mihir Sanjay and Love, Juliette and others},
  journal={arXiv preprint arXiv:2403.08295},
  year={2024}
}

@inproceedings{siglip,
  title={Sigmoid loss for language image pre-training},
  author={Zhai, Xiaohua and Mustafa, Basil and Kolesnikov, Alexander and Beyer, Lucas},
  booktitle={ICCV},
  year={2023}
}

@article{siglip2,
  title={Siglip 2: Multilingual vision-language encoders with improved semantic understanding, localization, and dense features},
  author={Tschannen, Michael and Gritsenko, Alexey and Wang, Xiao and Naeem, Muhammad Ferjad and Alabdulmohsin, Ibrahim and Parthasarathy, Nikhil and Evans, Talfan and Beyer, Lucas and Xia, Ye and Mustafa, Basil and others},
  journal={arXiv preprint arXiv:2502.14786},
  year={2025}
}

@inproceedings{cambrian,
  title={Cambrian-1: A fully open, vision-centric exploration of multimodal llms},
  author={Tong, Shengbang and Brown, Ellis and Wu, Penghao and Woo, Sanghyun and Middepogu, Manoj and Akula, Sai Charitha and Yang, Jihan and Yang, Shusheng and Iyer, Adithya and Pan, Xichen and others},
  booktitle={NeurIPS},
  year={2024}
}

@inproceedings{bert,
  title={Bert: Pre-training of deep bidirectional transformers for language understanding},
  author={Devlin, Jacob},
  booktitle={NAACL},
  year={2019}
}

@article{gpt3,
  title={Language models are few-shot learners},
  author={Brown, Tom B},
  journal={arXiv preprint arXiv:2005.14165},
  year={2020}
}

@article{coco,
  title={Microsoft coco captions: Data collection and evaluation server},
  author={Chen, Xinlei and Fang, Hao and Lin, Tsung-Yi and Vedantam, Ramakrishna and Gupta, Saurabh and Doll{\'a}r, Piotr and Zitnick, C Lawrence},
  journal={arXiv preprint arXiv:1504.00325},
  year={2015}
}

@article{llama,
  title={Llama: Open and efficient foundation language models},
  author={Touvron, Hugo and Lavril, Thibaut and Izacard, Gautier and Martinet, Xavier and Lachaux, Marie-Anne and Lacroix, Timoth{\'e}e and Rozi{\`e}re, Baptiste and Goyal, Naman and Hambro, Eric and Azhar, Faisal and others},
  journal={arXiv preprint arXiv:2302.13971},
  year={2023}
}

@inproceedings{imagenet,
  title={Imagenet: A large-scale hierarchical image database},
  author={Deng, Jia and Dong, Wei and Socher, Richard and Li, Li-Jia and Li, Kai and Fei-Fei, Li},
  booktitle={CVPR},
  year={2009},
}

@article{LAOIN,
  title={LAION-5B: An open large-scale dataset for training next generation image-text models},
  author={Christoph Schuhmann and Romain Beaumont and Richard Vencu and Cade Gordon and Ross Wightman and Mehdi Cherti and Theo Coombes and Aarush Katta and Clayton Mullis and Mitchell Wortsman and Patrick Schramowski and Srivatsa Kundurthy and Katherine Crowson and Ludwig Schmidt and Robert Kaczmarczyk and Jenia Jitsev},
  journal={ArXiv},
  year={2022},
  volume={abs/2210.08402},
  url={https://api.semanticscholar.org/CorpusID:252917726}
}

@article{metaclip2,
   title={MetaCLIP 2: A Worldwide Scaling Recipe},
   author={Yung-Sung Chuang and Yang Li and Dong Wang and Ching-Feng Yeh and Kehan Lyu and Ramya Raghavendra and James Glass and Lifei Huang and Jason Weston and Luke Zettlemoyer and Xinlei Chen and Zhuang Liu and Saining Xie and Wen-tau Yih and Shang-Wen Li and Hu Xu},
   journal={arXiv preprint arXiv:2507.22062},
   year={2025}
}

@article{metaclip,
   title={Demystifying CLIP Data},
   author={Hu Xu and Saining Xie and Xiaoqing Ellen Tan and Po-Yao Huang and Russell Howes and Vasu Sharma and Shang-Wen Li and Gargi Ghosh and Luke Zettlemoyer and Christoph Feichtenhofer},
   journal={arXiv preprint arXiv:2309.16671},
   year={2023}
}

@article{LLM_rewrite,
  title={Improving CLIP Training with Language Rewrites},
  author={Lijie Fan and Dilip Krishnan and Phillip Isola and Dina Katabi and Yonglong Tian},
  journal={ArXiv},
  year={2023},
  volume={abs/2305.20088},
  url={https://api.semanticscholar.org/CorpusID:258987272}
}

@article{Zhu2025InternVL3EA,
  title={InternVL3: Exploring Advanced Training and Test-Time Recipes for Open-Source Multimodal Models},
  author={Jinguo Zhu and Weiyun Wang and Zhe Chen and Zhaoyang Liu and Shenglong Ye and Lixin Gu and Yuchen Duan and Hao Tian and Weijie Su and Jie Shao and Zhangwei Gao and Erfei Cui and Yue Cao and Yangzhou Liu and Haomin Wang and Weiye Xu and Hao Li and Jiahao Wang and Han Lv and Dengnian Chen and Songze Li and Yinan He and Tan Jiang and Jiapeng Luo and Yi Wang and Cong He and Botian Shi and Xingcheng Zhang and Wenqi Shao and Junjun He and Ying Xiong and Wenwen Qu and Peng Sun and Penglong Jiao and Lijun Wu and Kai Zhang and Hui Deng and Jiaye Ge and Kaiming Chen and Limin Wang and Min Dou and Lewei Lu and Xizhou Zhu and Tong Lu and Dahua Lin and Yu Qiao and Jifeng Dai and Wenhai Wang},
  journal={ArXiv},
  year={2025},
  volume={abs/2504.10479},
  url={https://api.semanticscholar.org/CorpusID:277780955}
}

@inproceedings{simclr,
  title={A Simple Framework for Contrastive Learning of Visual Representations},
  author={Chen, Ting and Kornblith, Simon and Norouzi, Mohammad and Hinton, Geoffrey},
  booktitle={International Conference on Machine Learning},
  pages={1597--1607},
  year={2020},
  organization={PMLR}
}

@inproceedings{mae,
  title={Masked Autoencoders Are Scalable Vision Learners},
  author={He, Kaiming and Chen, Xin and Xie, Saining and Li, Yanghao and Doll{\'a}r, Piotr},
  booktitle={Proceedings of the IEEE/CVF Conference on Computer Vision and Pattern Recognition (CVPR)},
  pages={16000--16009},
  year={2022}
}

@inproceedings{dino,
  title={Emerging Properties in Self-Supervised Vision Transformers},
  author={Caron, Mathilde and Touvron, Hugo and Misra, Ishan and J{\'e}gou, Herv{\'e} and Mairal, Julien and Bojanowski, Piotr and Joulin, Armand},
  booktitle={Proceedings of the IEEE/CVF International Conference on Computer Vision (ICCV)},
  pages={9650--9660},
  year={2021}
}

@inproceedings{vit,
  title={An Image is Worth 16x16 Words: Transformers for Image Recognition at Scale},
  author={Dosovitskiy, Alexey and Beyer, Lucas and Kolesnikov, Alexander and Weissenborn, Dirk and Zhai, Xiaohua and Unterthiner, Thomas and Dehghani, Mostafa and Minderer, Matthias and Heigold, Georg and Gelly, Sylvain and Uszkoreit, Jakob and Houlsby, Neil},
  booktitle={International Conference on Learning Representations (ICLR)},
  year={2021}
}

@inproceedings{evit,
  title={EViT: Expediting Vision Transformers via Token Reorganizations},
  author={Youwei Liang and Chongjian Ge and Zhan Tong and Yibing Song and Jue Wang and Pengtao Xie},
  booktitle={International Conference on Learning Representations},
  year={2022},
  url={https://api.semanticscholar.org/CorpusID:251647803}
}

@inproceedings{singh2023effectiveness,
    title={The effectiveness of MAE pre-pretraining for billion-scale pretraining},
    author={Singh, Mannat and Duval, Quentin and Alwala, Kalyan Vasudev and Fan, Haoqi and Aggarwal, Vaibhav and Adcock, Aaron and Joulin, Armand and Doll{\'a}r, Piotr and Feichtenhofer, Christoph and Girshick, Ross and Girdhar, Rohit and Misra, Ishan},
    booktitle={ICCV},
    year={2023}
}

@article{Registers,
  title={Vision Transformers Need Registers},
  author={Timoth{\'e}e Darcet and Maxime Oquab and Julien Mairal and Piotr Bojanowski},
  journal={ArXiv},
  year={2023},
  volume={abs/2309.16588},
  url={https://api.semanticscholar.org/CorpusID:263134283}
}

@article{SLIP,
  title={SLIP: Self-supervision meets Language-Image Pre-training},
  author={Norman Mu and Alexander Kirillov and David A. Wagner and Saining Xie},
  journal={ArXiv},
  year={2021},
  volume={abs/2112.12750},
  url={https://api.semanticscholar.org/CorpusID:245424883}
}

@article{Mishra2022ASE,
  title={A simple, efficient and scalable contrastive masked autoencoder for learning visual representations},
  author={Shlok Kumar Mishra and Joshua Robinson and Huiwen Chang and David Jacobs and Aaron Sarna and Aaron Maschinot and Dilip Krishnan},
  journal={ArXiv},
  year={2022},
  volume={abs/2210.16870},
  url={https://api.semanticscholar.org/CorpusID:253237813}
}

@article{Oquab2023DINOv2LR,
  title={DINOv2: Learning Robust Visual Features without Supervision},
  author={Maxime Oquab and Timoth{\'e}e Darcet and Th{\'e}o Moutakanni and Huy Q. Vo and Marc Szafraniec and Vasil Khalidov and Pierre Fernandez and Daniel Haziza and Francisco Massa and Alaaeldin El-Nouby and Mahmoud Assran and Nicolas Ballas and Wojciech Galuba and Russ Howes and Po-Yao (Bernie) Huang and Shang-Wen Li and Ishan Misra and Michael G. Rabbat and Vasu Sharma and Gabriel Synnaeve and Huijiao Xu and Herv{\'e} J{\'e}gou and Julien Mairal and Patrick Labatut and Armand Joulin and Piotr Bojanowski},
  journal={ArXiv},
  year={2023},
  volume={abs/2304.07193},
  url={https://api.semanticscholar.org/CorpusID:258170077}
}

@article{Chen2023SymbolicDO,
  title={Symbolic Discovery of Optimization Algorithms},
  author={Xiangning Chen and Chen Liang and Da Huang and Esteban Real and Kaiyuan Wang and Yao Liu and Hieu Pham and Xuanyi Dong and Thang Luong and Cho-Jui Hsieh and Yifeng Lu and Quoc V. Le},
  journal={ArXiv},
  year={2023},
  volume={abs/2302.06675},
  url={https://api.semanticscholar.org/CorpusID:256846990}
}

@article{Shazeer2018AdafactorAL,
  title={Adafactor: Adaptive Learning Rates with Sublinear Memory Cost},
  author={Noam M. Shazeer and Mitchell Stern},
  journal={ArXiv},
  year={2018},
  volume={abs/1804.04235},
  url={https://api.semanticscholar.org/CorpusID:4786918}
}

@article{Bolya2022TokenMY,
  title={Token Merging: Your ViT But Faster},
  author={Daniel Bolya and Cheng-Yang Fu and Xiaoliang Dai and Peizhao Zhang and Christoph Feichtenhofer and Judy Hoffman},
  journal={ArXiv},
  year={2022},
  volume={abs/2210.09461},
  url={https://api.semanticscholar.org/CorpusID:252968113}
}

@article{Hu2021LoRALA,
  title={LoRA: Low-Rank Adaptation of Large Language Models},
  author={J. Edward Hu and Yelong Shen and Phillip Wallis and Zeyuan Allen-Zhu and Yuanzhi Li and Shean Wang and Weizhu Chen},
  journal={ArXiv},
  year={2021},
  volume={abs/2106.09685},
  url={https://api.semanticscholar.org/CorpusID:235458009}
}

@article{Huang2024LLM2CLIPPL,
  title={LLM2CLIP: Powerful Language Model Unlocks Richer Visual Representation},
  author={Weiquan Huang and Aoqi Wu and Yifan Yang and Xufang Luo and Yuqing Yang and Liang Hu and Qi Dai and Xiyang Dai and Dongdong Chen and Chong Luo and Lili Qiu},
  journal={ArXiv},
  year={2024},
  volume={abs/2411.04997},
  url={https://api.semanticscholar.org/CorpusID:273877534}
}

@article{Touvron2023LLaMAOA,
  title={LLaMA: Open and Efficient Foundation Language Models},
  author={Hugo Touvron and Thibaut Lavril and Gautier Izacard and Xavier Martinet and Marie-Anne Lachaux and Timoth{\'e}e Lacroix and Baptiste Rozi{\`e}re and Naman Goyal and Eric Hambro and Faisal Azhar and Aur'elien Rodriguez and Armand Joulin and Edouard Grave and Guillaume Lample},
  journal={ArXiv},
  year={2023},
  volume={abs/2302.13971},
  url={https://api.semanticscholar.org/CorpusID:257219404}
}

@article{Gao2021SimCSESC,
  title={SimCSE: Simple Contrastive Learning of Sentence Embeddings},
  author={Tianyu Gao and Xingcheng Yao and Danqi Chen},
  journal={ArXiv},
  year={2021},
  volume={abs/2104.08821},
  url={https://api.semanticscholar.org/CorpusID:233296292}
}

@article{bolya2025PerceptionEncoder,
  title={Perception Encoder: The best visual embeddings are not at the output of the network},
  author={Daniel Bolya and Po-Yao Huang and Peize Sun and Jang Hyun Cho and Andrea Madotto and Chen Wei and Tengyu Ma and Jiale Zhi and Jathushan Rajasegaran and Hanoona Rasheed and Junke Wang and Marco Monteiro and Hu Xu and Shiyu Dong and Nikhila Ravi and Daniel Li and Piotr Doll{\'a}r and Christoph Feichtenhofer},
  journal={arXiv:2504.13181},
  year={2025}
}

@misc{poolingattentioneffectivedesigns,
      title={Pooling And Attention: What Are Effective Designs For LLM-Based Embedding Models?}, 
      author={Yixuan Tang and Yi Yang},
      year={2024},
      eprint={2409.02727},
      archivePrefix={arXiv},
      primaryClass={cs.CL},
      url={https://arxiv.org/abs/2409.02727}, 
}

@article{Su2021RoFormerET,
  title={RoFormer: Enhanced Transformer with Rotary Position Embedding},
  author={Jianlin Su and Yu Lu and Shengfeng Pan and Bo Wen and Yunfeng Liu},
  journal={ArXiv},
  year={2021},
  volume={abs/2104.09864},
  url={https://api.semanticscholar.org/CorpusID:233307138}
}

@article{Zhai2023SigmoidLF,
  title={Sigmoid Loss for Language Image Pre-Training},
  author={Xiaohua Zhai and Basil Mustafa and Alexander Kolesnikov and Lucas Beyer},
  journal={2023 IEEE/CVF International Conference on Computer Vision (ICCV)},
  year={2023},
  pages={11941-11952},
  url={https://api.semanticscholar.org/CorpusID:257767223}
}

@article{Li2022ScalingLP,
  title={Scaling Language-Image Pre-Training via Masking},
  author={Yanghao Li and Haoqi Fan and Ronghang Hu and Christoph Feichtenhofer and Kaiming He},
  journal={2023 IEEE/CVF Conference on Computer Vision and Pattern Recognition (CVPR)},
  year={2022},
  pages={23390-23400},
  url={https://api.semanticscholar.org/CorpusID:254125280}
}

@article{Ghadiyaram2019LargeScaleWP,
  title={Large-Scale Weakly-Supervised Pre-Training for Video Action Recognition},
  author={Deepti Ghadiyaram and Matt Feiszli and Du Tran and Xueting Yan and Heng Wang and Dhruv Kumar Mahajan},
  journal={2019 IEEE/CVF Conference on Computer Vision and Pattern Recognition (CVPR)},
  year={2019},
  pages={12038-12047},
  url={https://api.semanticscholar.org/CorpusID:143423501}
}

@article{Singh2022RevisitingWS,
  title={Revisiting Weakly Supervised Pre-Training of Visual Perception Models},
  author={Mannat Singh and Laura Gustafson and Aaron B. Adcock and Vinicius de Freitas Reis and Buğra Gedik and Raj Prateek Kosaraju and Dhruv Kumar Mahajan and Ross B. Girshick and Piotr Doll'ar and Laurens van der Maaten},
  journal={2022 IEEE/CVF Conference on Computer Vision and Pattern Recognition (CVPR)},
  year={2022},
  pages={794-804},
  url={https://api.semanticscholar.org/CorpusID:246063491}
}

@inproceedings{Barbu2019ObjectNetAL,
  title={ObjectNet: A large-scale bias-controlled dataset for pushing the limits of object recognition models},
  author={Andrei Barbu and David Mayo and Julian Alverio and William Luo and Christopher Wang and Dan Gutfreund and Joshua B. Tenenbaum and Boris Katz},
  booktitle={Neural Information Processing Systems},
  year={2019},
  url={https://api.semanticscholar.org/CorpusID:202777185}
}

@article{Carreira2017QuoVA,
  title={Quo Vadis, Action Recognition? A New Model and the Kinetics Dataset},
  author={Jo{\~a}o Carreira and Andrew Zisserman},
  journal={2017 IEEE Conference on Computer Vision and Pattern Recognition (CVPR)},
  year={2017},
  pages={4724-4733},
  url={https://api.semanticscholar.org/CorpusID:206596127}
}

@article{Kuehne2011HMDBAL,
  title={HMDB: A large video database for human motion recognition},
  author={Hilde Kuehne and Hueihan Jhuang and Est{\'i}baliz Garrote and Tomaso A. Poggio and Thomas Serre},
  journal={2011 International Conference on Computer Vision},
  year={2011},
  pages={2556-2563},
  url={https://api.semanticscholar.org/CorpusID:206769852}
}

@article{Xu2016MSRVTTAL,
  title={MSR-VTT: A Large Video Description Dataset for Bridging Video and Language},
  author={Jun Xu and Tao Mei and Ting Yao and Yong Rui},
  journal={2016 IEEE Conference on Computer Vision and Pattern Recognition (CVPR)},
  year={2016},
  pages={5288-5296},
  url={https://api.semanticscholar.org/CorpusID:206594535}
}

@inproceedings{wang2019learning,
        title={Learning Robust Global Representations by Penalizing Local Predictive Power},
        author={Wang, Haohan and Ge, Songwei and Lipton, Zachary and Xing, Eric P},
        booktitle={Advances in Neural Information Processing Systems},
        pages={10506--10518},
        year={2019}
}

@article{chen2024deconstructing,
  title={Deconstructing denoising diffusion models for self-supervised learning},
  author={Chen, Xinlei and Liu, Zhuang and Xie, Saining and He, Kaiming},
  journal={arXiv preprint arXiv:2401.14404},
  year={2024}
}

@article{tschannen2024image,
  title={Image captioners are scalable vision learners too},
  author={Tschannen, Michael and Kumar, Manoj and Steiner, Andreas and Zhai, Xiaohua and Houlsby, Neil and Beyer, Lucas},
  journal={Advances in Neural Information Processing Systems},
  volume={36},
  year={2024}
}

@article{wan2024locca,
  title={LocCa: Visual Pretraining with Location-aware Captioners},
  author={Wan, Bo and Tschannen, Michael and Xian, Yongqin and Pavetic, Filip and Alabdulmohsin, Ibrahim and Wang, Xiao and Pinto, Andr{\'e} Susano and Steiner, Andreas and Beyer, Lucas and Zhai, Xiaohua},
  journal={arXiv preprint arXiv:2403.19596},
  year={2024}
}

@article{naeem2023silc,
  title={Silc: Improving vision language pretraining with self-distillation},
  author={Naeem, Muhammad Ferjad and Xian, Yongqin and Zhai, Xiaohua and Hoyer, Lukas and Van Gool, Luc and Tombari, Federico},
  journal={arXiv preprint arXiv:2310.13355},
  year={2023}
}

@inproceedings{dong2023maskclip,
  title={Maskclip: Masked self-distillation advances contrastive language-image pretraining},
  author={Dong, Xiaoyi and Bao, Jianmin and Zheng, Yinglin and Zhang, Ting and Chen, Dongdong and Yang, Hao and Zeng, Ming and Zhang, Weiming and Yuan, Lu and Chen, Dong and others},
  booktitle={Proceedings of the IEEE/CVF Conference on Computer Vision and Pattern Recognition},
  pages={10995--11005},
  year={2023}
}

@InProceedings{Sun_2024_CVPR,
    author    = {Sun, Zeyi and Fang, Ye and Wu, Tong and Zhang, Pan and Zang, Yuhang and Kong, Shu and Xiong, Yuanjun and Lin, Dahua and Wang, Jiaqi},
    title     = {Alpha-CLIP: A CLIP Model Focusing on Wherever You Want},
    booktitle = {Proceedings of the IEEE/CVF Conference on Computer Vision and Pattern Recognition (CVPR)},
    month     = {June},
    year      = {2024},
    pages     = {13019-13029}
}

@inproceedings{kirillov2023segment,
  title={Segment anything},
  author={Kirillov, Alexander and Mintun, Eric and Ravi, Nikhila and Mao, Hanzi and Rolland, Chloe and Gustafson, Laura and Xiao, Tete and Whitehead, Spencer and Berg, Alexander C and Lo, Wan-Yen and others},
  booktitle={Proceedings of the IEEE/CVF International Conference on Computer Vision},
  pages={4015--4026},
  year={2023}
}

@inproceedings{
chen2025contrastive,
title={Contrastive Localized Language-Image Pre-Training},
author={Hong-You Chen and Zhengfeng Lai and Haotian Zhang and Xinze Wang and Marcin Eichner and Keen You and Meng Cao and Bowen Zhang and Yinfei Yang and Zhe Gan},
booktitle={Forty-second International Conference on Machine Learning},
year={2025},
url={https://openreview.net/forum?id=sGQEOXlezg}
}

@inproceedings{fini2025multimodal,
  title={Multimodal autoregressive pre-training of large vision encoders},
  author={Fini, Enrico and Shukor, Mustafa and Li, Xiujun and Dufter, Philipp and Klein, Michal and Haldimann, David and Aitharaju, Sai and da Costa, Victor G Turrisi and B{\'e}thune, Louis and Gan, Zhe and others},
  booktitle={Proceedings of the Computer Vision and Pattern Recognition Conference},
  pages={9641--9654},
  year={2025}
}

@inproceedings{lai2024revisit,
  title={Revisit large-scale image-caption data in pre-training multimodal foundation models},
  author={Lai, Zhengfeng and Saveris, Vasileios and Chen, Chen and Chen, Hong-You and Zhang, Haotian and Zhang, Bowen and Tebar, Juan Lao and Hu, Wenze and Gan, Zhe and Grasch, Peter and others},
  booktitle={ICLR},
  year={2025}
}

@article{bica2024improving,
  title={Improving fine-grained understanding in image-text pre-training},
  author={Bica, Ioana and Ili{\'c}, Anastasija and Bauer, Matthias and Erdogan, Goker and Bo{\v{s}}njak, Matko and Kaplanis, Christos and Gritsenko, Alexey A and Minderer, Matthias and Blundell, Charles and Pascanu, Razvan and others},
  journal={arXiv preprint arXiv:2401.09865},
  year={2024}
}

@article{nguyen2023improving,
  title={Improving multimodal datasets with image captioning},
  author={Nguyen, Thao and Gadre, Samir Yitzhak and Ilharco, Gabriel and Oh, Sewoong and Schmidt, Ludwig},
  journal={Advances in neural information processing systems},
  volume={36},
  pages={22047--22069},
  year={2023}
}

@inproceedings{jia2021scaling,
  title={Scaling up visual and vision-language representation learning with noisy text supervision},
  author={Jia, Chao and Yang, Yinfei and Xia, Ye and Chen, Yi-Ting and Parekh, Zarana and Pham, Hieu and Le, Quoc and Sung, Yun-Hsuan and Li, Zhen and Duerig, Tom},
  booktitle={International conference on machine learning},
  pages={4904--4916},
  year={2021},
  organization={PMLR}
}

@inproceedings{xie2022simmim,
  title={Simmim: A simple framework for masked image modeling},
  author={Xie, Zhenda and Zhang, Zheng and Cao, Yue and Lin, Yutong and Bao, Jianmin and Yao, Zhuliang and Dai, Qi and Hu, Han},
  booktitle=cvpr,
  pages={9653--9663},
  year={2022}
}

@inproceedings{bao2021beit,
  title={{BE}i{T}: {BERT} pre-training of image transformers},
  author={Bao, Hangbo and Dong, Li and Wei, Furu},
  booktitle=iclr,
  year={2022}
}

@inproceedings{chen2022context,
  title={Context autoencoder for self-supervised representation learning},
  author={Chen, Xiaokang and Ding, Mingyu and Wang, Xiaodi and Xin, Ying and Mo, Shentong and Wang, Yunhao and Han, Shumin and Luo, Ping and Zeng, Gang and Wang, Jingdong},
  booktitle={preprint arXiv:2202.03026},
  year={2022}
}

@inproceedings{tian2020contrastive,
  title={Contrastive multiview coding},
  author={Tian, Yonglong and Krishnan, Dilip and Isola, Phillip},
  booktitle=eccv,
  pages={776--794},
  year={2020},
}

@article{oord2018representation,
  title={Representation learning with contrastive predictive coding},
  author={van den Oord, A{\"a}ron and Li, Yazhe and Vinyals, Oriol},
  journal={preprint arXiv:1807.03748},
  year={2018}
}

@inproceedings{he2020momentum,
  title={Momentum contrast for unsupervised visual representation learning},
  author={He, Kaiming and Fan, Haoqi and Wu, Yuxin and Xie, Saining and Girshick, Ross},
  booktitle=cvpr,
  pages={9729--9738},
  year={2020}
}

@inproceedings{devlin2018bert,
  title={{BERT}: Pre-training of deep bidirectional transformers for language understanding},
  author={Devlin, Jacob and Chang, Ming-Wei and Lee, Kenton and Toutanova, Kristina},
  booktitle={NAACL},
  year={2018}
}

@inproceedings{chen2021empirical,
  title={An empirical study of training self-supervised vision transformers},
  author={Chen, Xinlei and Xie, Saining and He, Kaiming},
  booktitle=iccv,
  pages={9640--9649},
  year={2021}
}

@InProceedings{robinson2020contrastive,
  title={Contrastive learning with hard negative samples},
  author={Robinson, Joshua and Chuang, Ching-Yao and Sra, Suvrit and Jegelka, Stefanie},
  booktitle=iclr,
  year={2021}
}

@InProceedings{chuang2020debiased,
  title={Debiased contrastive learning},
  author={Chuang, Ching-Yao and Robinson, Joshua and Lin, Yen-Chen and Torralba, Antonio and Jegelka, Stefanie},
  booktitle={NEURIPS},
  volume={33},
  pages={8765--8775},
  year={2020}
}

@inproceedings{radford2021learning,
  title={Learning transferable visual models from natural language supervision},
  author={Radford, Alec and Kim, Jong Wook and Hallacy, Chris and Ramesh, Aditya and Goh, Gabriel and Agarwal, Sandhini and Sastry, Girish and Askell, Amanda and Mishkin, Pamela and Clark, Jack and others},
  booktitle={International Conference on Machine Learning},
  pages={8748--8763},
  year={2021},
  organization={PMLR}
}

@inproceedings{henaff2020data,
  title={Data-efficient image recognition with contrastive predictive coding},
  author={H{\'e}naff, Olivier and Srinivas, Aravind and De Fauw, Jeffrey and Razavi, Ali and Doersch, Carl and Eslami, S. M. Ali and van den Oord, A{\"a}ron},
  booktitle={International Conference on Machine Learning (ICML)},
  pages={4182--4192},
  year={2020}
}

@inproceedings{Ge2021RobustCL,
  title={Robust Contrastive Learning Using Negative Samples with Diminished Semantics},
  author={Songwei Ge and Shlok Kumar Mishra and Haohan Wang and Chun-Liang Li and David Jacobs},
  booktitle={NEURIPS},
  year={2021},
  volume={abs/2110.14189}
}

@misc{he2015deepresiduallearningimage,
      title={Deep Residual Learning for Image Recognition}, 
      author={Kaiming He and Xiangyu Zhang and Shaoqing Ren and Jian Sun},
      year={2015},
      eprint={1512.03385},
      archivePrefix={arXiv},
      primaryClass={cs.CV},
      url={https://arxiv.org/abs/1512.03385}, 
}

@misc{kay2017kineticshumanactionvideo,
      title={The Kinetics Human Action Video Dataset}, 
      author={Will Kay and Joao Carreira and Karen Simonyan and Brian Zhang and Chloe Hillier and Sudheendra Vijayanarasimhan and Fabio Viola and Tim Green and Trevor Back and Paul Natsev and Mustafa Suleyman and Andrew Zisserman},
      year={2017},
      eprint={1705.06950},
      archivePrefix={arXiv},
      primaryClass={cs.CV},
      url={https://arxiv.org/abs/1705.06950}, 
}

@misc{grattafiori2024llama3herdmodels,
      title={The Llama 3 Herd of Models}, 
      author={Aaron Grattafiori and Abhimanyu Dubey and Abhinav Jauhri and Abhinav Pandey and Abhishek Kadian and Ahmad Al-Dahle and Aiesha Letman and Akhil Mathur and Alan Schelten and Alex Vaughan and Amy Yang and Angela Fan and Anirudh Goyal and Anthony Hartshorn and Aobo Yang and Archi Mitra and Archie Sravankumar and Artem Korenev and Arthur Hinsvark and Arun Rao and Aston Zhang and Aurelien Rodriguez and Austen Gregerson and Ava Spataru and Baptiste Roziere and Bethany Biron and Binh Tang and Bobbie Chern and Charlotte Caucheteux and Chaya Nayak and Chloe Bi and Chris Marra and Chris McConnell and Christian Keller and Christophe Touret and Chunyang Wu and Corinne Wong and Cristian Canton Ferrer and Cyrus Nikolaidis and Damien Allonsius and Daniel Song and Danielle Pintz and Danny Livshits and Danny Wyatt and David Esiobu and Dhruv Choudhary and Dhruv Mahajan and Diego Garcia-Olano and Diego Perino and Dieuwke Hupkes and Egor Lakomkin and Ehab AlBadawy and Elina Lobanova and Emily Dinan and Eric Michael Smith and Filip Radenovic and Francisco Guzmán and Frank Zhang and Gabriel Synnaeve and Gabrielle Lee and Georgia Lewis Anderson and Govind Thattai and Graeme Nail and Gregoire Mialon and Guan Pang and Guillem Cucurell and Hailey Nguyen and Hannah Korevaar and Hu Xu and Hugo Touvron and Iliyan Zarov and Imanol Arrieta Ibarra and Isabel Kloumann and Ishan Misra and Ivan Evtimov and Jack Zhang and Jade Copet and Jaewon Lee and Jan Geffert and Jana Vranes and Jason Park and Jay Mahadeokar and Jeet Shah and Jelmer van der Linde and Jennifer Billock and Jenny Hong and Jenya Lee and Jeremy Fu and Jianfeng Chi and Jianyu Huang and Jiawen Liu and Jie Wang and Jiecao Yu and Joanna Bitton and Joe Spisak and Jongsoo Park and Joseph Rocca and Joshua Johnstun and Joshua Saxe and Junteng Jia and Kalyan Vasuden Alwala and Karthik Prasad and Kartikeya Upasani and Kate Plawiak and Ke Li and Kenneth Heafield and Kevin Stone and Khalid El-Arini and Krithika Iyer and Kshitiz Malik and Kuenley Chiu and Kunal Bhalla and Kushal Lakhotia and Lauren Rantala-Yeary and Laurens van der Maaten and Lawrence Chen and Liang Tan and Liz Jenkins and Louis Martin and Lovish Madaan and Lubo Malo and Lukas Blecher and Lukas Landzaat and Luke de Oliveira and Madeline Muzzi and Mahesh Pasupuleti and Mannat Singh and Manohar Paluri and Marcin Kardas and Maria Tsimpoukelli and Mathew Oldham and Mathieu Rita and Maya Pavlova and Melanie Kambadur and Mike Lewis and Min Si and Mitesh Kumar Singh and Mona Hassan and Naman Goyal and Narjes Torabi and Nikolay Bashlykov and Nikolay Bogoychev and Niladri Chatterji and Ning Zhang and Olivier Duchenne and Onur Çelebi and Patrick Alrassy and Pengchuan Zhang and Pengwei Li and Petar Vasic and Peter Weng and Prajjwal Bhargava and Pratik Dubal and Praveen Krishnan and Punit Singh Koura and Puxin Xu and Qing He and Qingxiao Dong and Ragavan Srinivasan and Raj Ganapathy and Ramon Calderer and Ricardo Silveira Cabral and Robert Stojnic and Roberta Raileanu and Rohan Maheswari and Rohit Girdhar and Rohit Patel and Romain Sauvestre and Ronnie Polidoro and Roshan Sumbaly and Ross Taylor and Ruan Silva and Rui Hou and Rui Wang and Saghar Hosseini and Sahana Chennabasappa and Sanjay Singh and Sean Bell and Seohyun Sonia Kim and Sergey Edunov and Shaoliang Nie and Sharan Narang and Sharath Raparthy and Sheng Shen and Shengye Wan and Shruti Bhosale and Shun Zhang and Simon Vandenhende and Soumya Batra and Spencer Whitman and Sten Sootla and Stephane Collot and Suchin Gururangan and Sydney Borodinsky and Tamar Herman and Tara Fowler and Tarek Sheasha and Thomas Georgiou and Thomas Scialom and Tobias Speckbacher and Todor Mihaylov and Tong Xiao and Ujjwal Karn and Vedanuj Goswami and Vibhor Gupta and Vignesh Ramanathan and Viktor Kerkez and Vincent Gonguet and Virginie Do and Vish Vogeti and Vítor Albiero and Vladan Petrovic and Weiwei Chu and Wenhan Xiong and Wenyin Fu and Whitney Meers and Xavier Martinet and Xiaodong Wang and Xiaofang Wang and Xiaoqing Ellen Tan and Xide Xia and Xinfeng Xie and Xuchao Jia and Xuewei Wang and Yaelle Goldschlag and Yashesh Gaur and Yasmine Babaei and Yi Wen and Yiwen Song and Yuchen Zhang and Yue Li and Yuning Mao and Zacharie Delpierre Coudert and Zheng Yan and Zhengxing Chen and Zoe Papakipos and Aaditya Singh and Aayushi Srivastava and Abha Jain and Adam Kelsey and Adam Shajnfeld and Adithya Gangidi and Adolfo Victoria and Ahuva Goldstand and Ajay Menon and Ajay Sharma and Alex Boesenberg and Alexei Baevski and Allie Feinstein and Amanda Kallet and Amit Sangani and Amos Teo and Anam Yunus and Andrei Lupu and Andres Alvarado and Andrew Caples and Andrew Gu and Andrew Ho and Andrew Poulton and Andrew Ryan and Ankit Ramchandani and Annie Dong and Annie Franco and Anuj Goyal and Aparajita Saraf and Arkabandhu Chowdhury and Ashley Gabriel and Ashwin Bharambe and Assaf Eisenman and Azadeh Yazdan and Beau James and Ben Maurer and Benjamin Leonhardi and Bernie Huang and Beth Loyd and Beto De Paola and Bhargavi Paranjape and Bing Liu and Bo Wu and Boyu Ni and Braden Hancock and Bram Wasti and Brandon Spence and Brani Stojkovic and Brian Gamido and Britt Montalvo and Carl Parker and Carly Burton and Catalina Mejia and Ce Liu and Changhan Wang and Changkyu Kim and Chao Zhou and Chester Hu and Ching-Hsiang Chu and Chris Cai and Chris Tindal and Christoph Feichtenhofer and Cynthia Gao and Damon Civin and Dana Beaty and Daniel Kreymer and Daniel Li and David Adkins and David Xu and Davide Testuggine and Delia David and Devi Parikh and Diana Liskovich and Didem Foss and Dingkang Wang and Duc Le and Dustin Holland and Edward Dowling and Eissa Jamil and Elaine Montgomery and Eleonora Presani and Emily Hahn and Emily Wood and Eric-Tuan Le and Erik Brinkman and Esteban Arcaute and Evan Dunbar and Evan Smothers and Fei Sun and Felix Kreuk and Feng Tian and Filippos Kokkinos and Firat Ozgenel and Francesco Caggioni and Frank Kanayet and Frank Seide and Gabriela Medina Florez and Gabriella Schwarz and Gada Badeer and Georgia Swee and Gil Halpern and Grant Herman and Grigory Sizov and Guangyi and Zhang and Guna Lakshminarayanan and Hakan Inan and Hamid Shojanazeri and Han Zou and Hannah Wang and Hanwen Zha and Haroun Habeeb and Harrison Rudolph and Helen Suk and Henry Aspegren and Hunter Goldman and Hongyuan Zhan and Ibrahim Damlaj and Igor Molybog and Igor Tufanov and Ilias Leontiadis and Irina-Elena Veliche and Itai Gat and Jake Weissman and James Geboski and James Kohli and Janice Lam and Japhet Asher and Jean-Baptiste Gaya and Jeff Marcus and Jeff Tang and Jennifer Chan and Jenny Zhen and Jeremy Reizenstein and Jeremy Teboul and Jessica Zhong and Jian Jin and Jingyi Yang and Joe Cummings and Jon Carvill and Jon Shepard and Jonathan McPhie and Jonathan Torres and Josh Ginsburg and Junjie Wang and Kai Wu and Kam Hou U and Karan Saxena and Kartikay Khandelwal and Katayoun Zand and Kathy Matosich and Kaushik Veeraraghavan and Kelly Michelena and Keqian Li and Kiran Jagadeesh and Kun Huang and Kunal Chawla and Kyle Huang and Lailin Chen and Lakshya Garg and Lavender A and Leandro Silva and Lee Bell and Lei Zhang and Liangpeng Guo and Licheng Yu and Liron Moshkovich and Luca Wehrstedt and Madian Khabsa and Manav Avalani and Manish Bhatt and Martynas Mankus and Matan Hasson and Matthew Lennie and Matthias Reso and Maxim Groshev and Maxim Naumov and Maya Lathi and Meghan Keneally and Miao Liu and Michael L. Seltzer and Michal Valko and Michelle Restrepo and Mihir Patel and Mik Vyatskov and Mikayel Samvelyan and Mike Clark and Mike Macey and Mike Wang and Miquel Jubert Hermoso and Mo Metanat and Mohammad Rastegari and Munish Bansal and Nandhini Santhanam and Natascha Parks and Natasha White and Navyata Bawa and Nayan Singhal and Nick Egebo and Nicolas Usunier and Nikhil Mehta and Nikolay Pavlovich Laptev and Ning Dong and Norman Cheng and Oleg Chernoguz and Olivia Hart and Omkar Salpekar and Ozlem Kalinli and Parkin Kent and Parth Parekh and Paul Saab and Pavan Balaji and Pedro Rittner and Philip Bontrager and Pierre Roux and Piotr Dollar and Polina Zvyagina and Prashant Ratanchandani and Pritish Yuvraj and Qian Liang and Rachad Alao and Rachel Rodriguez and Rafi Ayub and Raghotham Murthy and Raghu Nayani and Rahul Mitra and Rangaprabhu Parthasarathy and Raymond Li and Rebekkah Hogan and Robin Battey and Rocky Wang and Russ Howes and Ruty Rinott and Sachin Mehta and Sachin Siby and Sai Jayesh Bondu and Samyak Datta and Sara Chugh and Sara Hunt and Sargun Dhillon and Sasha Sidorov and Satadru Pan and Saurabh Mahajan and Saurabh Verma and Seiji Yamamoto and Sharadh Ramaswamy and Shaun Lindsay and Shaun Lindsay and Sheng Feng and Shenghao Lin and Shengxin Cindy Zha and Shishir Patil and Shiva Shankar and Shuqiang Zhang and Shuqiang Zhang and Sinong Wang and Sneha Agarwal and Soji Sajuyigbe and Soumith Chintala and Stephanie Max and Stephen Chen and Steve Kehoe and Steve Satterfield and Sudarshan Govindaprasad and Sumit Gupta and Summer Deng and Sungmin Cho and Sunny Virk and Suraj Subramanian and Sy Choudhury and Sydney Goldman and Tal Remez and Tamar Glaser and Tamara Best and Thilo Koehler and Thomas Robinson and Tianhe Li and Tianjun Zhang and Tim Matthews and Timothy Chou and Tzook Shaked and Varun Vontimitta and Victoria Ajayi and Victoria Montanez and Vijai Mohan and Vinay Satish Kumar and Vishal Mangla and Vlad Ionescu and Vlad Poenaru and Vlad Tiberiu Mihailescu and Vladimir Ivanov and Wei Li and Wenchen Wang and Wenwen Jiang and Wes Bouaziz and Will Constable and Xiaocheng Tang and Xiaojian Wu and Xiaolan Wang and Xilun Wu and Xinbo Gao and Yaniv Kleinman and Yanjun Chen and Ye Hu and Ye Jia and Ye Qi and Yenda Li and Yilin Zhang and Ying Zhang and Yossi Adi and Youngjin Nam and Yu and Wang and Yu Zhao and Yuchen Hao and Yundi Qian and Yunlu Li and Yuzi He and Zach Rait and Zachary DeVito and Zef Rosnbrick and Zhaoduo Wen and Zhenyu Yang and Zhiwei Zhao and Zhiyu Ma},
      year={2024},
      eprint={2407.21783},
      archivePrefix={arXiv},
      primaryClass={cs.AI},
      url={https://arxiv.org/abs/2407.21783}, 
}

@misc{assran2025vjepa2selfsupervisedvideo,
      title={V-JEPA 2: Self-Supervised Video Models Enable Understanding, Prediction and Planning}, 
      author={Mido Assran and Adrien Bardes and David Fan and Quentin Garrido and Russell Howes and Mojtaba and Komeili and Matthew Muckley and Ammar Rizvi and Claire Roberts and Koustuv Sinha and Artem Zholus and Sergio Arnaud and Abha Gejji and Ada Martin and Francois Robert Hogan and Daniel Dugas and Piotr Bojanowski and Vasil Khalidov and Patrick Labatut and Francisco Massa and Marc Szafraniec and Kapil Krishnakumar and Yong Li and Xiaodong Ma and Sarath Chandar and Franziska Meier and Yann LeCun and Michael Rabbat and Nicolas Ballas},
      year={2025},
      eprint={2506.09985},
      archivePrefix={arXiv},
      primaryClass={cs.AI},
      url={https://arxiv.org/abs/2506.09985}, 
}

@misc{fan2023improvingcliptraininglanguage,
      title={Improving CLIP Training with Language Rewrites}, 
      author={Lijie Fan and Dilip Krishnan and Phillip Isola and Dina Katabi and Yonglong Tian},
      year={2023},
      eprint={2305.20088},
      archivePrefix={arXiv},
      primaryClass={cs.CV},
      url={https://arxiv.org/abs/2305.20088}, 
}

@misc{tran2015learningspatiotemporalfeatures3d,
      title={Learning Spatiotemporal Features with 3D Convolutional Networks}, 
      author={Du Tran and Lubomir Bourdev and Rob Fergus and Lorenzo Torresani and Manohar Paluri},
      year={2015},
      eprint={1412.0767},
      archivePrefix={arXiv},
      primaryClass={cs.CV},
      url={https://arxiv.org/abs/1412.0767}, 
}

@misc{siméoni2025dinov3,
      title={DINOv3}, 
      author={Oriane Siméoni and Huy V. Vo and Maximilian Seitzer and Federico Baldassarre and Maxime Oquab and Cijo Jose and Vasil Khalidov and Marc Szafraniec and Seungeun Yi and Michaël Ramamonjisoa and Francisco Massa and Daniel Haziza and Luca Wehrstedt and Jianyuan Wang and Timothée Darcet and Théo Moutakanni and Leonel Sentana and Claire Roberts and Andrea Vedaldi and Jamie Tolan and John Brandt and Camille Couprie and Julien Mairal and Hervé Jégou and Patrick Labatut and Piotr Bojanowski},
      year={2025},
      eprint={2508.10104},
      archivePrefix={arXiv},
      primaryClass={cs.CV},
      url={https://arxiv.org/abs/2508.10104}, 
}

@misc{feichtenhofer2022maskedautoencodersspatiotemporallearners,
      title={Masked Autoencoders As Spatiotemporal Learners}, 
      author={Christoph Feichtenhofer and Haoqi Fan and Yanghao Li and Kaiming He},
      year={2022},
      eprint={2205.09113},
      archivePrefix={arXiv},
      primaryClass={cs.CV},
      url={https://arxiv.org/abs/2205.09113}, 
}

@misc{mahajan2018exploringlimitsweaklysupervised,
      title={Exploring the Limits of Weakly Supervised Pretraining}, 
      author={Dhruv Mahajan and Ross Girshick and Vignesh Ramanathan and Kaiming He and Manohar Paluri and Yixuan Li and Ashwin Bharambe and Laurens van der Maaten},
      year={2018},
      eprint={1805.00932},
      archivePrefix={arXiv},
      primaryClass={cs.CV},
      url={https://arxiv.org/abs/1805.00932}, 
}

\appendix 

\section{Model Architecture (Detailed)}
\begin{figure}[ht]
    \centering
    \includegraphics[width=\textwidth]{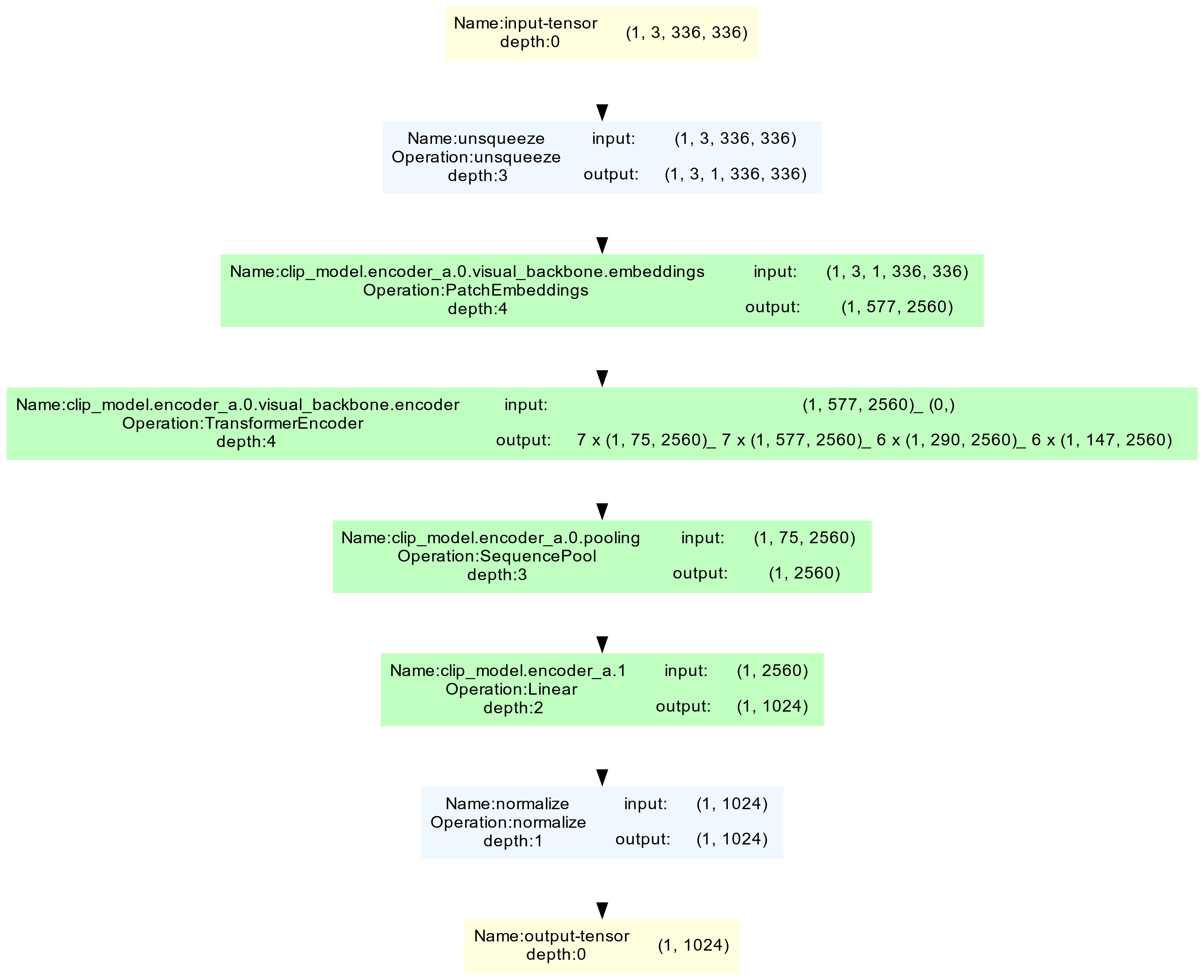}
    \caption{XrayVisual Model Overview}
    \label{fig:ap:xrv}
\end{figure}


\end{document}